\documentclass[10pt]{article}
\usepackage{graphicx}
\usepackage{graphics}
\usepackage{subcaption}
\usepackage{amsmath}
\usepackage[ansinew]{inputenc}
\usepackage{amsfonts}
\usepackage{bm}
\usepackage{float}
\usepackage{url}
\usepackage[round, authoryear]{natbib}
\usepackage{color}
\usepackage{multirow}

\DeclareMathOperator{\erf}{erf} 
 \DeclareMathOperator{\tr}{tr}
\DeclareMathOperator{\cov}{cov} \DeclareMathOperator{\ex}{E}

\DeclareMathOperator{\entropy}{H}

\newcommand{\dif}{\textrm{d}}

\newcommand{\boldgamma}{\bm{\gamma}}
\newcommand{\boldZ}{\mathbf{Z}}
\newcommand{\boldS}{\mathbf{S}}

\newcommand{\boldy}{\mathbf{y}} 
\newcommand{\boldx}{\mathbf{x}} 
\newcommand{\boldz}{\mathbf{z}} 
\newcommand{\boldk}{\mathbf{k}} 
\newcommand{\boldK}{\mathbf{K}} 
\newcommand{\boldf}{\mathbf{f}} 
\newcommand{\boldP}{\mathbf{P}} 
\newcommand{\boldA}{\mathbf{A}} 
\newcommand{\boldu}{\mathbf{u}} 
\newcommand{\boldupsi}{\bm{\upsilon}}
\newcommand{\boldX}{\mathbf{X}} 
\newcommand{\inputSpace}{\mathcal{X}} 
\newcommand{\params}{\bm{\theta}} 
\newcommand{\gauss}{\mathcal{N}} 
\newcommand{\gammad}{\operatorname{Gamma}} 
\newcommand{\betad}{\operatorname{Beta}} 
\newcommand{\bernd}{\operatorname{Bernoulli}} 
\newcommand{\ones}{\mathbf{1}}

\title{Indian Buffet process for model selection in convolved
  multiple-output Gaussian processes}

\author{Cristian Guarnizo, Mauricio A. \'Alvarez\\ {\small
    \emph{Faculty of Engineering, Universidad Tecnol\'ogica de
      Pereira, Pereira, Colombia.}}}

\date{}

\begin{document}
\maketitle

\begin{abstract}
  Multi-output Gaussian processes have received increasing attention
  during the last few years as a natural mechanism to extend the
  powerful flexibility of Gaussian processes to the setup of multiple
  output variables. The key point here is the ability to design kernel
  functions that allow exploiting the correlations between the outputs
  while fulfilling the positive definiteness requisite for the
  covariance function. Alternatives to construct these covariance
  functions are the linear model of coregionalization and process
  convolutions. Each of these methods demand the specification of the
  number of latent Gaussian process used to build the covariance
  function for the outputs. We propose in this paper, the use of an
  Indian Buffet process as a way to perform model selection over the
  number of latent Gaussian processes. This type of model is
  particularly important in the context of latent force models, where
  the latent forces are associated to physical quantities like protein
  profiles or latent forces in mechanical systems. We use variational
  inference to estimate posterior distributions over the variables
  involved, and show examples of the model performance over artificial
  data, a motion capture dataset, and a gene expression dataset.
\end{abstract}

\section{Introduction}
Kernel methods for vector-valued functions have proved to be an
important tool for designing learning algorithms that perform
multi-variate regression \citep{Bonilla:multi07}, and multi-class
classification \citep{Skolidis:multiclassICM2011}. A kernel function
that encodes suitable correlations between output variables can be
embedded in established machine learning algorithms like support
vector machines or Gaussian process predictors, where the kernel
function is interpreted as a covariance function.

Different kernels for vector-valued functions proposed in recent years
within the machine learning community, are particular cases of the so
called linear model of coregionalization (LMC)
\citep{Journel:miningBook78, Goovaerts:book97}, heavily used for
cokriging in geostatistics \citep{Chiles:book99,
  Cressie:spatialdataBook:1993}. Furthermore, the linear model of
coregionalization turns out to be a special case of the so called
process convolutions (PC) used in statistics for developing covariance
functions \citep{verHoef:convolution98,Higdon:ocean98}. For details,
see \citet{Alvarez:review:2012}.

Under the LMC or the PC frameworks, the way in which a kernel function
for multiple variables is constructed, follows a similar pattern: a
set of orthogonal Gaussian processes, each of them characterized by
an specific covariance function, are linearly combined to represent each
of the output variables. Typically, each of these Gaussian processes
establishes the degree of smoothness that is to be explained in the
outputs. In PC, the set of orthogonal Gaussian processes are initially
smoothed through a convolution operation that involves the
specification of the so called smoothing kernels.  The smoothing
kernel may be the impulse response of a dynamical system, or, in
general, may correspond to the Green's function associated to a
differential equation. Gaussian processes that use a kernel
constructed from a PC with a Green's function as a smoothing kernel,
have been coined by the authors of \citet{Alvarez:lfm09} as latent
force models.

Despite its success for prediction, it is still unclear how to select
the number of orthogonal Gaussian processes used for building the
multi-output Gaussian process or a latent force model. Furthermore, in
the context of latent force models where these orthogonal Gaussian
processes may represent a physical quantity, like the action of a
protein for transcription regulation of a gene or a latent force in a
system involving masses and dampers, it becomes relevant to undercover
the interactions between the latent Gaussian processes and the output
variables that are being modelled.

In this paper, we use an Indian Buffet Process (IBP)
\citep{Griffiths05infinitelatent, Griffiths:IBPjmlr:2011} for model
selection in convolved multiple output Gaussian processes.  The IBP is
a non-parametric prior over binary matrices, that imposes an structure
over the sparsity pattern of the binary matrix. It has previously been
used for introducing sparsity in linear models
\citep{Knowles:annals:2011}. We formulate a variational inference
procedure for inferring posterior distributions over the structure of
the relationships between output functions and latent processes, by
combining ideas from \citet{Alvarez:inducing10} and
\citet{DoshiVelez:VIBP:2009}. We show examples of the model using
artificial data, motion capture data and a gene expression
dataset.

\section{Convolved multiple output Gaussian processes}
We want to jointly model $D$ output functions
$\{f_d(\boldx)\}_{d=1}^D$, where each output $f_d(\boldx)$ can be
written as
\begin{align}\label{lfm1}
f_d(\boldx) & = \sum_{q=1}^Q S_{d,q}\int_{\mathcal{X}}G_{d}(\boldx-\boldx')u_q(\boldx')\operatorname{d}\boldx',
\end{align}
where $G_{d}(\boldx-\boldx')$ are smoothing functions or smoothing kernels,
$\{u_q(\boldx)\}_{q=1}^Q$ are orthogonal processes, and the variables $\{S_{d,q}\}_{d=1, q=1}^{D,Q}$ measure the influence of the latent function $q$ over the output function $d$. We assume that each latent process $u_q(\boldx)$ is a Gaussian process with zero mean function and covariance function $k_q(\boldx, \boldx')$.

The model above is known in the geostatistics literature as a process
convolution. If $G_{d}(\cdot)$ is equal to the Dirac delta function, then the linear model of coregionalization is recovered \citep{Alvarez:review:2012}. Also, in the context of linear dynamical systems, the function $G_d(\cdot)$ is related to the so called impulse response of the system.

\subsection{Covariance functions}

Due to the linearity in expression \eqref{lfm1}, the set of processes
$\{f_d(\boldx)\}_{d=1}^D$ follow a joint Gaussian process with mean
function equal to zero, and covariance function given by
\begin{align*}
k_{f_d, f_{d'}}(\boldx, \boldx') & =
\operatorname{cov}\left[f_d(\boldx)f_{d'}(\boldx')\right] =
\sum_{q=1}^Q S_{d,q}S_{d',q}k_{f^q_d,f^q_{d'}}(\boldx, \boldx'),
\end{align*}
where we have defined
\begin{align}\label{eq:kfdqkfdq}
k_{f^q_d,f^q_{d'}}(\boldx, \boldx')=\int_{\mathcal{X}}\int_{\mathcal{X}}
  G_{d}(\boldx-\boldz)G_{d'}(\boldx-\boldz')k_q(\boldz,\boldz')
  \operatorname{d}\boldz\operatorname{d}\boldz'.
\end{align}
Besides the covariance function defined above, we are interested in the covariance function between $f_d(\boldx)$ and $u_q(\boldx)$, which follows
\begin{align}\label{eq:kfdqkuq}
   k_{f_d, u_q}(\boldx, \boldx')&=
\operatorname{cov}\left[f_d(\boldx)u_{q}(\boldx')\right] = S_{d,q}\int_{\mathcal{X}}G_{d}(\boldx-\boldz)k_q(\boldz,\boldx')
  \operatorname{d}\boldz.
\end{align}

For some forms of the smoothing kernel $G_d(\cdot)$, and the covariance function $k_q(\cdot)$, the covariance functions $k_{f^q_d, f^q_{d'}}(\boldx, \boldx')$ and $k_{f_d,u_q}(\boldx, \boldx')$ can be worked out analytically. We show some examples in section \ref{s:covfun}.

\subsection{Likelihood model for multi-output regression}

In a multi-variate regression setting the likelihood model for each output
can be expressed as
\begin{align*}
  y_d(\boldx) = f_d(\boldx) + w_d(\boldx),
\end{align*}
where each $f_d(\boldx)$ is given by \eqref{lfm1}, and
$\{w_d(\boldx)\}_{d=1}^D$ are a set of processes that
could represent a noise process for each output. Assuming that each
$w_d(\boldx)$ is also a Gaussian process with zero mean and covariance
function $k_{w_d, w_{d}}(\boldx,\boldx')$, the covariance function between
$y_d(\boldx)$, and $y_{d'}(\boldx')$ is given by
\begin{align*}
 k_{y_d, y_{d'}}(\boldx, \boldx') &= k_{f_d, f_{d'}}(\boldx, \boldx')
 + k_{w_d, w_{d}}(\boldx, \boldx')\delta_{d,d'},
\end{align*}
where $\delta_{d,d'}$ is the Kronecker delta.

\subsection{Inference and hyper-parameter learning}

Let $\mathcal{D} = \{\boldX_d, \boldy_d\}_{d=1}^D$ be a dataset for a
multi-output regression problem. We use $\boldX$ to refer to the set
$\{\boldX_d\}_{d=1}^D$, and $\boldy$ to refer to the set
$\{\boldy_d\}_{d=1}^D$. We assume that we have $N$ data observations
for each output. The posterior distribution $p(\boldu|\boldy)$, and
the predictive distribution for $\boldf_*$ at a new input point
$\boldx_*$ can both be computed using standard Gaussian processes
formulae \citep{Rasmussen:book06}.

Different methods have been proposed for performing computationally efficient inference and hyperparameter learning in multi-output Gaussian processes \citep{Alvarez:JMLR:2011}, reducing the computational complexity from $\mathcal{O}(N^3D^3)$ to $\mathcal{O}(NDM^2)$, where $M$ is a user-specified value.

\section{The Indian Buffet process}
An open question in models like the one described in Eq. \eqref{lfm1} is how to choose the number of latent functions $Q$. In this report, we will use an Indian Buffet process as a prior to automatically choose $Q$.

The IBP is a distribution over binary matrices with a finite number of
rows and an unbounded number of columns \citep{Griffiths05infinitelatent}. This can define a non-parametric latent feature model in which rows are related to data points and columns are related to latent features. The relationship between latent features and data points can be encoded in a binary matrix $\boldZ \in
\mathbb{R}^{D\times Q}$, where $Z_{d,q}=1$ if feature $q$ is used to
explain data point $d$ and $Z_{d,q}=0$ otherwise. Each element $Z_{d,q}$ of the matrix $\boldZ$ is sampled as follows
\begin{align*}
v_j    &\sim \text{Beta}(\alpha,1),\\
 \pi_q &= \prod_{j=1}^{q} v_j,\\
Z_{d,q} & \sim \text{Bernoulli}(\pi_q),
\end{align*}
where $\alpha$ is a real positive value, and $\pi_q$ is the probability of observing a non-zero value in the column $q$ of the matrix $\boldZ$, this is, the value $\pi_q$ controls the sparsity for 
the latent feature $q$. As we will see, in our proposed model, the value of $\alpha$ is related to the average number of latent functions
per output.

Using an IBP as a prior for a linear Gaussian model, the authors in \citet{DoshiVelez:VIBP:2009}, derive two variational mean field approximations, referred to as a ``finite variational approach'', and an ``infinite variational approach''. We adopt the latter approach, this is, even though that the update equations will be based on the true IBP posterior over an infinite number of features, for a practical implementation, we use a level of truncation $Q$ as the maximum number of latent functions. In this approach, as shown in previous equations, $\boldupsi=\{v_j\}_{j=1}^Q $ are independent samples from a Beta distribution, while $\bm{\pi}=\{\pi_q\}_{q=1}^Q$ are dependent variables obtained by multiplying the sampled values for $v_j$ as shown before. Thus, in the factorised variational distribution of our mean field approach we use $\boldupsi$ as hidden variables with the prior given before. We induce sparsity over the sensitivities by pre-multiplying $Z_{d,q}$ with $S_{d,q}$, as explained in the next section.
\section{Variational formulation for model selection}\label{model}
The model selection approach presented here is based on the variational formulation for convolved multiple output Gaussian processes proposed by \citet{Alvarez:inducing10}, and the variational formulation for the Indian Buffet Process proposed by \citet{DoshiVelez:VIBP:2009}. We start by defining the likelihood as
\begin{align}
p(\mathbf{y}|u,\mathbf{X},\bm{\theta}, \boldS, \boldZ)& = \gauss(\boldy|\boldf,\bm{\Sigma}_{\mathbf{w}}) \nonumber
=\prod_{d=1}^D
\gauss(\boldy_d|\boldf_d,\bm{\Sigma}_{\mathbf{w}_d}),
\end{align}
where $u=\{u_q\}_{q=1}^Q$, $\boldS=[S_{d,q}]\in \Re^{D\times Q}$,
$\boldZ=[Z_{d,q}]\in \{0,1\}^{D\times Q}$ and each output vector
$\boldf_d$ is defined as
\begin{align*}
\begin{bmatrix}
  f_d(\boldx_1) \\
  f_d(\boldx_2) \\
  \vdots \\
  f_d(\boldx_N)
\end{bmatrix}
=
\begin{bmatrix}
  \sum_{q=1}^QZ_{d,q}S_{d,q}\int_{\inputSpace}G_{d,q}(\boldx_1-\boldz)u_{q}(\boldz)\dif \boldz\\
   \sum_{q=1}^QZ_{d,q}S_{d,q}\int_{\inputSpace}G_{d,q}(\boldx_2-\boldz)u_{q}(\boldz)\dif \boldz\\
  \vdots \\
   \sum_{q=1}^QZ_{d,q}S_{d,q}\int_{\inputSpace}G_{d,q}(\boldx_N-\boldz)u_{q}(\boldz)\dif \boldz
\end{bmatrix}.
\end{align*}
For each latent function $u_q(\cdot)$, we define a set of auxiliary variables or inducing variables $\boldu_q \in \mathbb{R}$, obtained when evaluating the latent function $u_q$ at a set of $M$ inducing inputs $\{\boldz_m\}_{m=1}^M$. We refer to the set of inducing variables using $\boldu = \{\boldu_q\}_{q=1}^Q$. Following ideas used in several computationally efficient Gaussian process methods, we work with the conditional densities $p(u|\boldu)$, instead of the full Gaussian process $p(u)$. The conditional density of the latent functions given the inducing variables can be written as
\begin{align*}
p(u|\boldu) = & \prod_{q=1}^Q\gauss(u_q|\boldk^{\top}_{u_q,\boldu_q}\boldK_{\boldu_q,\boldu_q}^{-1}\boldu_q,k_{u_q,u_q} - \boldk^{\top}_{u_q,\boldu_q}\boldK_{\boldu_q,\boldu_q}^{-1}\boldk_{u_q,\boldu_q}),
\end{align*}
with $\boldk^{\top}_{u_q,\boldu_q}=[k_{u_q,u_q}(\boldz,\boldz_1),
  k_{u_q,u_q}(\boldz,\boldz_2), \ldots,
  k_{u_q,u_q}(\boldz,\boldz_M)]$.
The prior over $\boldu$ has the following form
\begin{equation*}
p(\boldu) = \prod_{q=1}^Q\gauss(\boldu_q|\mathbf{0}, \boldK_{\boldu_q,\boldu_q}).
\end{equation*}
For the elements of $\boldS$ we use an spike and slab prior as follows
\citep{Knowles:annals:2011}
\begin{align*}
p(S_{d,q}|Z_{d,q}, \gamma_{d,q}) = & Z_{d,q}\gauss(S_{d,q}|0, \gamma^{-1}_{d,q})  + (1-Z_{d,q})\delta(S_{d,q}),
\end{align*}
where $Z_{d,q}$ are the elements of the binary matrix $\boldZ$ that follows an Indian Buffet Process Prior. This is different from \cite{Titsias:SpikeSlab:2011} where all the variables $Z_{d,q}$ are drawn from the same Bernoulli distribution with parameter $\pi$.  From the previous section we know that the prior for $Z_{d,q}$ is given by
\begin{align*}
  p(Z_{d,q}|\pi_q)& = \bernd(Z_{d,q}|\pi_q).
\end{align*}
To apply the variational method, we write the joint distribution for $Z_{d,q}$ and $S_{d,q}$ as
\begin{align*}
  p(S_{d,q}, Z_{d,q}|\pi_q, \gamma_{d,q}) =& p(S_{d,q}|Z_{d,q}, \gamma_{d,q})p(Z_{d,q}|\pi_q) \nonumber\\
= & [\pi_q\gauss(S_{d,q}|0, \gamma^{-1}_{d,q})]^{Z_{d,q}} [(1- \pi_q)\delta(S_{d,q})]^{1-Z_{d,q}},
\end{align*}
leading to
\begin{align*}
p(\boldS, \boldZ|\boldupsi, \boldgamma) = & \prod_{d=1}^D\prod_{q=1}^Q [\pi_q\gauss(S_{d,q}|0, \gamma^{-1}_{d,q})]^{Z_{d,q}} [(1- \pi_q)\delta(S_{d,q})]^{1-Z_{d,q}}.
\end{align*}
We assume that the hyperparameters $\gamma_{d,q}$ follow a Gamma prior,
\begin{align*}
p(\boldgamma)& = \prod_{d=1}^D\prod_{q=1}^Q\gammad(\gamma_{d,q}|a^{\gamma}_{d,q},b^{\gamma}_{d,q}).
\end{align*}
Although not written explicitly, the idea here is that $Q$ can be as
high as we want to. In fact, the IBP prior assumes that $Q\rightarrow\infty$, but in our variational inference implementations the value of $Q$ is fixed and it represents the truncation level on the IBP  (see \citet{DoshiVelez:VIBP:2009}). According to our model, the complete likelihood follows as
\begin{align*}
p(\boldy, \boldX, u, \boldS, \boldZ, \boldupsi, \boldgamma, \params)  = &
p(\boldy|\boldX, u, \boldS, \boldZ, \boldupsi, \boldgamma, \params)p(u|\params) p(\boldS, \boldZ|\boldgamma, \boldupsi)
p(\boldupsi)p(\boldgamma),
\end{align*}
where $\params$ are the hyperparameters regarding the type of covariance function (see section \ref{s:covfun}). For the variational distribution, we use a mean field approximation, and assume that the terms in the posterior factorize as
\begin{align*}
  q(\boldu) =  \prod_{q=1}^Qq(\boldu_q),
  \quad
  q(\boldS, \boldZ) = \prod_{d=1}^D\prod_{q=1}^Q q(S_{d,q}| Z_{d,q})q(Z_{d,q}),
  \quad
  q(\boldgamma)
  = \prod_{d=1}^D\prod_{q=1}^Q q(\gamma_{d,q}),
  \quad 
  q(\boldupsi) = \prod_{q=1}^Q q(\upsilon_q).
\end{align*}
Following the same formulation used by \citet{Alvarez:VIKs:report:2009}, the posterior takes the form
\begin{align*}
q(u,\mathbf{u}, \boldS,\boldZ, \boldupsi,\boldgamma)&=
p(u|\mathbf{u})q(\mathbf{u})q(\boldS,\boldZ)q(\boldgamma)q(\boldupsi).
\end{align*}
The lower bound that needs to be maximized, $F_{V}(q(\mathbf{u}),q(\boldS, \boldZ),q(\boldgamma),q(\boldupsi))$, is given as \citep{Bishop:PRLM06}
\begin{align*}
& \int
q(u,\mathbf{u}, \boldS,\boldZ, \boldupsi,\boldgamma)
\log\left\{\frac{p(\boldy|\boldX, u, \boldS, \boldZ, \boldupsi, \boldgamma, \params)
p(u|\mathbf{u})p(\mathbf{u})p(\boldS,\boldZ|\boldupsi, \boldgamma)
p(\boldupsi)p(\boldgamma)}
{q(u,\mathbf{u}, \boldS,\boldZ, \boldupsi,\boldgamma)}\right\}\dif{u}\,\dif{\mathbf{u}}\,\dif{\boldS}\, \dif{\boldZ}\, \dif{\boldupsi}\,\dif{\boldgamma}.
\end{align*}
By using standard variational equations \citep{Bishop:PRLM06}, it can be shown that the lower bound $F_V$ is given as
\begin{align*}
F_{V} &= \frac{1}{2}\boldy^{\top}\bm{\Sigma}_{\mathbf{w}}^{-1} \widetilde{\boldK}_{\boldf, \boldu}
\widetilde{\boldA}^{-1}\widetilde{\boldK}_{\boldf, \boldu}^{\top}\bm{\Sigma}_{\mathbf{w}}^{-1}\boldy
-\frac{1}{2}\log |\widetilde{\boldA}|+\frac{1}{2} \log |\boldK_{\boldu,\boldu}|
-\frac{1}{2}\log |\bm{\Sigma}| - \frac{1}{2}\tr(\bm{\Sigma}_{\mathbf{w}}^{-1}\boldy\boldy^{\top})\\
&-\frac{1}{2}\sum_{d=1}^D\sum_{q=1}^Q \ex[Z_{d,q}S_{d,q}^2] \tr\left[\mathbf{\Sigma}^{-1}_{\mathbf{w}_d}
\boldK_{\boldf_d|\boldu_q}\right] - \frac{1}{2}\log 2\pi\sum_{d=1}^D\sum_{q=1}^Q\ex[Z_{d,q}] + \sum_{d=1}^D\sum_{q=1}^Q \ex[Z_{d,q}]\ex[\log \pi_q] \\
& +
\frac{1}{2}\sum_{d=1}^D\sum_{q=1}^Q\ex[Z_{d,q}]\left[\psi(a^{\gamma^*}_{d,q})-\log b^{\gamma^*}_{d,q}\right]
-\frac{1}{2}\sum_{d=1}^D\sum_{q=1}^Q\frac{a^{\gamma^*}_{d,q}}{b^{\gamma^*}_{d,q}}\ex[Z_{d,q}S_{d,q}^2] \\
&+ \sum_{d=1}^D\sum_{q=1}^Q(1-\ex[Z_{d,q}])\ex[\log(1- \pi_q)]
-\sum_{d=1}^D\sum_{q=1}^Q\log\Gamma(a^{\gamma}_{d,q}) + \sum_{d=1}^D\sum_{q=1}^Q a^{\gamma}_{d,q}\log b^{\gamma}_{d,q}\\
&+ \sum_{d=1}^D\sum_{q=1}^Q\left(a^{\gamma}_{d,q}-1\right)\left[\psi(a^{\gamma^*}_{d,q}) -\log b^{\gamma^*}_{d,q}\right]
-\sum_{d=1}^D\sum_{q=1}^Q b^{\gamma}_{d,q}\frac{a^{\gamma^*}_{d,q}}{b^{\gamma^*}_{d,q}}
+ (\alpha - 1)\sum_{q=1}^Q\left[ \psi(\tau_{q1}) - \psi(\tau_{q1} + \tau_{q2}) \right]\\
&+\sum_{d=1}^D\sum_{q=1}^Q\eta_{d,q}
\left[\frac{1}{2} \log v_{d,q} +\frac{1}{2}(1 +\log(2\pi))\right]
+ \sum_{d=1}^D\sum_{q=1}^Q [-\eta_{d,q} \log \eta_{d,q} - (1-\eta_{d,q})\log(1-\eta_{d,q})]\\
&+\sum_{q=1}^Q\left[\log\left(\frac{\Gamma(\tau_{q1})\Gamma(\tau_{q2})}{\Gamma(\tau_{q1} + \tau_{q2})}
\right)
-(\tau_{q1} -1 )\psi(\tau_{q1}) -(\tau_{q2} -1 )\psi(\tau_{q2}) + (\tau_{q1} + \tau_{q2} -2 )\psi(\tau_{q1}+ \tau_{q2})\right] \\
&+  \sum_{d=1}^D\sum_{q=1}^Q\left[\log \Gamma(a^{\gamma^*}_{d,q}) - (a^{\gamma^*}_{d,q}
- 1)\psi(a^{\gamma^*}_{d,q}) - \log b^{\gamma^*}_{d,q} +a^{\gamma^*}_{d,q}\right],
\end{align*}
where $\widetilde{\boldK}_{\boldf, \boldu} =
\mathbf{E}_{\boldZ\boldS}\odot \boldK_{\boldf, \boldu}$ and
$\widetilde{\boldA} = \boldA + \widetilde{\boldK}^{\top}_{\boldf,
  \boldu}\bm{\Sigma}_{\mathbf{w}}^{-1}\widetilde{\boldK}_{\boldf,
  \boldu}$. Besides, $\mathbf{E}_{\boldZ\boldS}$ is a block-wise matrix with blocks $\ex[Z_{d,q}S_{d,q}]\mathbf{1}_{N\times N}$, $\boldA = \boldK_{\boldu, \boldu} + \overline{\boldK}_{\boldu, \boldu}$, $\overline{\boldK}_{\boldu, \boldu}= \mathbf{M}\odot
\left(\widehat{\boldK}^{\top}_{\boldf,   \boldu}\bm{\Sigma}_{\mathbf{w}}^{-1} \widehat{\boldK}_{\boldf,
  \boldu}\right)$, with $\mathbf{M}$ being a block-diagonal matrix
with blocks $\ones_{N\times N}$, and $\widehat{\boldK}_{\boldf,
  \boldu}=\mathbf{V}_{\boldZ\boldS}\odot \boldK_{\boldf, \boldu}$,
with $\mathbf{V}_{\boldZ\boldS}$ being a block-wise matrix with blocks
given by $\left(\ex[Z_{d,q}S^2_{d,q}] - \ex[Z_{d,q}S_{d,q}]^2\right)\ones_{N\times N}$. The operator $\odot$ refers to an element-wise product, and it is also known as the Hadamard product. While, $\psi(\cdot)$ and $\Gamma(\cdot)$ are the digamma and gamma function, respectively. It is important to notice that the first six terms of the lower bound defined above have the same form as the lower bound found in \citet{Alvarez:inducing10, Titsias:2009:VarInd}.

To perform variational inference, we obtain the update equations for the posterior distributions $q(\boldu)$, $q(\boldS, \boldZ)$, $q(\boldgamma)$, and $q(\boldupsi)$ from the lower bound given above. The update equations are included in appendix \ref{UpDist}, while the mean and the variance for the predictive distribution $p(\boldy_*|\boldy,\bm{\theta})$ appear on appendix \ref{PreDist}.

\section{Related work}
Convolved multiple output Gaussian Processes have been successfully
applied to regression tasks, such as motion capture data, gene
expression data, sensor network data, among others. In the context of
latent force models, multiple output Gaussian processes can be used to
probabilistically describe several interconnected dynamical systems,
with the advantage that the differential equations that describe those
systems, and the data observations, work together for accomplishing
system identification \citep{Alvarez:LFM:PAMI:2013}. Multiple output
Gaussian processes are commonly trained assuming that each output is
fully connected to all latent functions, this is, the value of each
hidden variable $Z_{d,q}$ is equal to one for all $d$, and $q$.

Several methods have been proposed in the literature for the problem
of model selection in related areas of multiple output Gaussian
processes. For example in multi-task learning, a Bayesian multi-task
learning model capable to learn the sparsity pattern of the data
features base on matrix-variate Gaussian scale mixtures is proposed in
\citet{Guo:NIPS:2011}. Later, a multi-task learning algorithm that
allows sharing one or more latent basis for task belonging to
different groups is presented in \citet{Kumar:ICML:2012}. This
algorithm is also capable of finding the number of latent basis, but
it does not place a matrix variate prior ever the sensitivities.

In a closely related work in multi-task Gaussian processes
\citep{Titsias:SpikeSlab:2011}, the problem of model selection was
approached using the spike and slab distribution as prior over the
weight matrix of a linear combination of Gaussian processes latent
functions. The inference step is performed using the variational approach.

\section{Implementation}
In this section, we briefly describe the covariance functions used in
the experiments. The first type of covariance function is based on a
convolution of two exponential functions with squared argument. The
second type of covariance functions is based on the solution of
ordinary differentials equations (ODE), and each data point is linked
to a time value.

\subsection{Covariance functions}\label{s:covfun}
We use three different types of kernels derived from expressions
$k_{f_d^q, f_{d'}^q}(\boldx,\boldx')$ in \eqref{eq:kfdqkfdq}, and
$k_{f_d^q, u_q}(\boldx,\boldx')$ in \eqref{eq:kfdqkuq}. In turn, these
expressions depend of the particular forms for $G_d(\cdot)$, and
$k_q(\cdot,\cdot)$.

\subsubsection{General purpose covariance function}\label{gs}
Here we present a general purpose covariance function for multi-output
GPs for which $\boldx\in \mathcal{X}\in \mathbb{R}^p$.
If we assume that both the smoothing kernel $G_d(\cdot)$ and $k_q(\cdot,\cdot)$ have the following form
\begin{align*}
k(\boldx,\boldx') & =  \frac{|\boldP|^{1/2}}{(2\pi)^{p/2}}\exp\left[
-\frac{1}{2} (\boldx - \boldx')^{\top}\boldP (\boldx - \boldx') \right],
\end{align*}
where $\boldP$ is a precision matrix, then it can be shown that
the covariance function
$k_{f_d^q, f_{d'}^q}(\boldx,\boldx')$ follows as
\begin{align*}
k_{f_d^q,f_{d'}^{q}}(\boldx,\boldx') =
\frac{1}{(2\pi)^{p/2}|\boldP^q_{d,d'}|^{1/2}}\exp\left[ -\frac{1}{2}
  (\boldx - \boldx')^{\top}(\boldP^q_{d,d'})^{-1} (\boldx - \boldx')
  \right],
\end{align*}
where $\boldP^q_{d,d'} = \boldP_d^{-1} + \boldP_{d'}^{-1} +
\bm{\Lambda}_q^{-1}$. Matrices $\boldP_d$ and $\bm{\Lambda}_q$
correspond to the precision matrices associated to $G_d(\cdot)$, and
$k_q(\cdot, \cdot)$, respectively. For the experiments, we use
diagonal forms for both matrices, where $\{p_{d,i}\}_{i=1}^p$ are the
elements for the diagonal matrix $\boldP_d$, and
$\{\ell_{q,i}\}_{i=1}^p$ are the diagonal elements of the matrix
$\bm{\Lambda}_q$. In the following sections, we refer to this covariance
function as the Gaussian Smoothing (GS) kernel.
\subsubsection{Latent force models}\label{lfm}
Latent force models (LFM) can be seen as a hybrid approach that
combines differential equations and Gaussian processes \citep{Alvarez:lfm09, Alvarez:LFM:PAMI:2013}.  They are built from
convolution processes by means of a deterministic function (which
relates the data to a physical model), and Gaussian process priors for
the latent functions. In the next two sections we show examples with a
first order ordinary differential equation (ODE1), and a second order
ordinary differential equation (ODE2). In both cases, we assume the latent functions to be Gaussian processes with zero mean and covariance functions given by
\begin{align}\label{Kuu1d}
k_q(t,t') = \exp\left[- \frac{(t-t')^2}{l_q^2} \right]. 
\end{align}
We can derive the covariance function for the outputs following
\eqref{eq:kfdqkfdq}. In this context, the smooothig kernel
$G_d(\cdot)$ is known as the Green's function, and its form will
depend on the order of the differential equation.
\paragraph{First order differential equation (ODE1)}\label{sec:ode1}
We assume that the data can be modelled by the first order differential equation given by
\begin{align} \label{fos} 
\frac{\text{d}f_d(t)}{\text{d}t}+B_df_d(t) & = \sum\limits_{q=1}^{Q}
S_{d,q} u_q(t),
\end{align}
where $B_d$ is the decay rate for output $d$. Solving for $f_d(t)$ in Equation \eqref{fos}, we get a similar expression to the one obtained in equation \eqref{lfm1}, where the smoothing kernel $G_d(\cdot)$ (or the Green's function in this context) is given by
\begin{align*} 
G_{d}(t-t') & = \exp\left[ - B_d(t-t') \right]. 
\end{align*}
Using the above form for the smoothing kernel $G_d(\cdot)$, and the
covariance function $k_q(\cdot, \cdot)$ given in \eqref{Kuu1d}, we derive the expression for $k_{f^q_d,f^q_{d'}}(t, t')$ using \eqref{eq:kfdqkfdq}, which is \citep{Lawrence:NIPS:2006}
\begin{align*}
k_{f^q_d,f^q_{d'}} &= \frac{\sqrt{\pi}l_q}{2}[ h_q(B_{d'},B_d,t',t) + h_q(B_{d},B_{d'},t,t')],
\end{align*}
where $h_q(B_{d'},B_d,t',t)$ is defined as
\begin{align}\label{hq} h_q(B_{d'},B_{d},t',t) =& \frac{\exp(\nu^2_{q,d'})}{B_d+B_{d'}} \exp(-B_{d'}t') \biggl\{ \exp(B_{d'}t) \left[ \erf\left(\frac{t'-t}{l_q} - \nu_{q,d'} \right) + \erf\left( \frac{t}{l_q} + \nu_{q,d'} \right) \right] \nonumber \\& -\exp(-B_dt) \left[ \erf\left( \frac{t'}{l_q}  - \nu_{q,d'}\right) + \erf(\nu_{q,d'}) \right] \biggr\},
\end{align}
with $\nu_{q,d} = l_qB_d/2$, and $\erf(\cdot)$ is the error function defined as
\[ \erf(z) = \frac{1}{2\pi} \int_{0}^{z} \exp\left( -t^2 \right) \text{d}t.
\]
In order to infer the latent functions $u_q(t)$ related in
\eqref{fos}, we calculate the cross-covariance function between
$f_d(t)$ and $u_q(t)$ using \eqref{eq:kfdqkuq}, as
follows \begin{align*} k_{f_d, u_q}(t,t') =& S_{d,q}
  \frac{\sqrt{\pi}l_q}{2} \exp(\nu_{q,d}^2)\exp\left[- B_d(t-t')
    \right]\left[ \erf \left(\frac{t-t'}{l_q} - \nu_{q,d} \right) +
    \erf \left(\frac{t'}{l_q} + \nu_{q,d} \right) \right].
\end{align*}

\paragraph{Second order differential equation (ODE2)}
In this scenario, we assume that the data can be explained using a second order differential equation related to a mechanical system
\begin{equation} \label{sos}
  m_d\frac{\text{d}^2f_d(t)}{\text{d}t^2} + C_d \frac{\text{d}f_d(t)}{\text{d}t}
  +B_df_d(t) = \sum\limits_{q=1}^{Q} S_{d,q} u_q(t),
\end{equation}
where $\{m_d\}_{d=1}^D$ are mass constants, $\{C_d\}_{d=1}^D$ are
damper constants, and $\{B_d\}_{d=1}^D$ are spring constants. Without loss of generality, the value of the mass $m_d$ is set to one. Now, assuming initial conditions equal to zero, the solution for the Green's function associated to \eqref{sos} is given by 
\[ G_d(t-t') = \frac{1}{\omega_d}\exp\left[ - \alpha_d(t-t') \right]\sin\left[\omega_d(t-t') \right],
\]
where $\alpha_d$ is the decay rate and $\omega_d$ is the natural
frequency. Both variables are defined as
\[
\alpha_d= \frac{C_d}{2}, \quad \quad\omega_d = \frac{\sqrt{4 B_d -
    C_d^2}}{2}.
\]
It can be shown that $k_{f_d^q,f_{d'}^q}(t,t')$ reduces to \citep{Alvarez:lfm09}
\begin{align*}
k_{f_d^q,f_{d'}^q}(t,t') = &
K_0[h_q(\tilde{\gamma}_{d'},\gamma_d,t,t') +
  h_q(\gamma_{d},\tilde{\gamma}_{d'},t',t) +
  h_q(\gamma_{d'},\tilde{\gamma}_{d},t,t') +
  h_q(\tilde{\gamma}_d,\gamma_{d'},t',t) \nonumber\\&
  -h_q(\tilde{\gamma}_{d'},\tilde{\gamma}_{d},t,t') -
  h_q(\tilde{\gamma}_{d},\tilde{\gamma}_{d'},t',t) -
  h_q(\tilde{\gamma}_{d'},\tilde{\gamma}_{d},t,t') -
  h_q(\gamma_{d},\gamma_{d'},t',t)],
\end{align*}
where $K_0 = \frac{l_q \sqrt{\pi}}{8\omega_d \omega_{d'}}$
$\gamma_d= \alpha_d+j\omega_d$, $\tilde{\gamma}_d= \alpha_d-j\omega_d$ and $h_q(\cdot)$ is the function defined in \eqref{hq}. Additionally, if $\omega_d$ and $\omega_{d'}$ take real values, the expression above simplifies as
\begin{align*}
k_{f_d^q,f_{d'}^q}(t,t') =& 2K_0 \operatorname{Re}\left[h_q(\gamma_{d'},\tilde{\gamma}_d,t,t') + h_q(\tilde{\gamma_{d}},\gamma_{d'},t',t) - h_q(\gamma_{d'},\gamma_{d},t,t') - h_q(\gamma_{d},\gamma_{d'},t',t) \right],
\end{align*}
where $\operatorname{Re}(\cdot)$ refers to the real part of the argument. For the cross-covariance $k_{f_d,u_q}(t,t')$, it can be shown that the solution for \eqref{eq:kfdqkuq} is
\[ k_{f_d,u_q}(t,t') = \frac{l_qS_{d,q}\sqrt{\pi}}{j4\omega_d}[\Upsilon_q(\tilde{\gamma}_d,t,t')-\Upsilon_q(\gamma_d,t,t')],
\]
where
\[ \Upsilon_q(\gamma_{d},t,t') = e^{\frac{l_q^2 \gamma_{d}^2}{4}}e^{-\gamma_{d}(t-t')}\left[  \erf\left(\frac{t-t'}{l_q} - \frac{l_q\gamma_{d}}{2}\right) + \erf\left( \frac{t'}{l_q} + \frac{l_q\gamma_{d}}{2} \right) \right].
\]
\subsection{Variational inference procedure}
The variational inference procedure can be summarized as follows. We
give initial values to the parameters of each variational
distribution, and initial values to the parameters of the covariance
functions. We also set the values of $\alpha$, and $Q$. An
iterative process is then performed until a criterion of convergence is
fulfilled. At each iteration, we update the moments for each
variational distribution as shown in appendix \ref{UpDist}. Alongside,
every ten iterations in the variational inference method, we estimate
the parameters $\params$ of the kernel functions, by maximizing the
lower bound $F_V$ using the scaled conjugate gradient method.
The derivatives  $\frac{\partial F_{V}}{\partial \boldK_{\boldu,\boldu}}$,
$\frac{\partial F_{V}}{\partial \boldK_{\boldu,\boldf}}$ and $\frac{\partial F_{V}}{\partial \boldK_{\boldf,\boldf}}$, are calculated using expressions similar to the ones obtained in \citet{Alvarez:VIKs:report:2009}.
We combine those derivatives with the
derivatives of $\boldK_{\boldu,\boldu}$, $\boldK_{\boldu,\boldf}$, and  $\boldK_{\boldf,\boldf}$ wrt $\params$.
We use the software GPmat (\url{https://github.com/SheffieldML/GPmat}) to train and test models based on latent force models.

\section{Results}
In this section, we show results from different datasets, including:
artificial data, motion capture data, and gene expression data. For the
artificial datasets, we are interested in recovering the known
interconnection matrix ($\boldZ$) between the latent functions and
outputs. For the real datasets, we analyse the regression performance of the
proposed method under different configurations.

\subsection{Synthetic Data} 
To show the ability of the proposed model to recover the underlying
structure between the output data and the latent functions, we apply the method to two different toy multi-output datasets. Each toy dataset is built by sampling from the model explained in section \ref{model}.

\paragraph{Example 1:}
The first experiment is conducted using a GS covariance function (see section \ref{gs}) and sample from the model with $D = 3$, $Q=2$ and $\alpha =1 $. For the smoothing kernels $G_d(x, x')$, we set the length-scales to $p_{1,1}= 0.01$, $p_{2,1} = 1/120$, and $p_{3,1} = 1/140$. We use the following values for matrices $\boldZ$, and $\boldS$,
\[ \boldZ = \left[ \begin{array}{cc} 0 & 1\\ 1 & 0\\ 1 & 0 \end{array} \right], \quad \boldS = \left[ \begin{array}{cc} 0 & 1.48\\ -3.19 & 0\\ 6.87 & 0 \end{array} \right]
\]
For the covariance functions $k_q(x,x')$ of the latent functions, 
we choose the length-scales as $l_{1,1}=0.1$ and $l_{2,1} = 0.2$.
Next, we sample the model and generate 30 data points per output, evenly spaced in the interval $[0,1]$. We assume that each process $w_d(x)$ is a white Gaussian noise process with zero mean, and
standard deviation equal to 0.1.

The model is then trained using the proposed variational method with a maximum number of latent functions set to four. Additionally, for the variational distribution of latent functions, we set 15 inducing points evenly space along the output interval.
\begin{figure}[h]
        \centering
        \begin{subfigure}[b]{0.22\textwidth}
        \includegraphics[width=\textwidth]{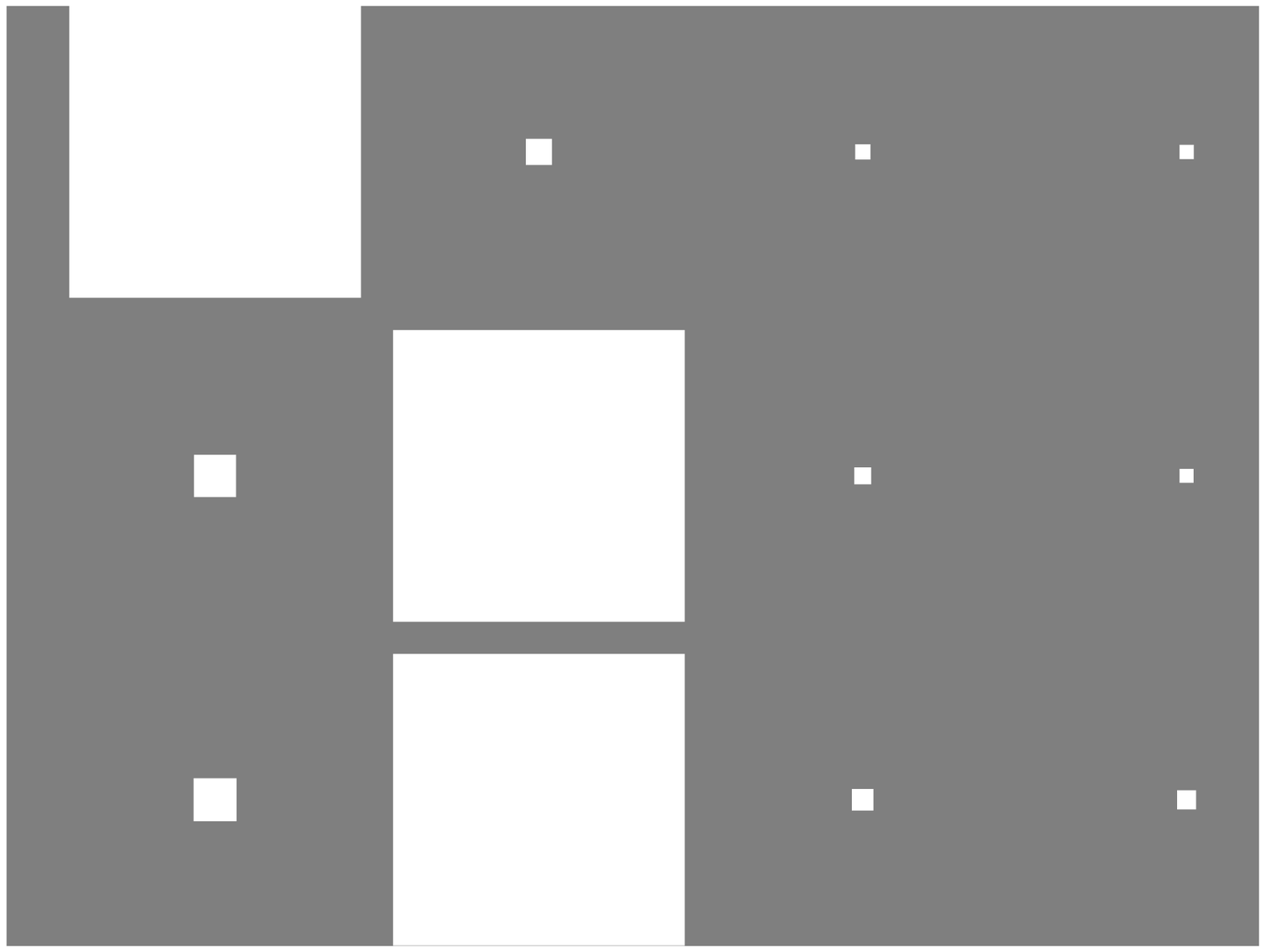}
        \caption{Hinton diagram toy 1}
        \label{Hinton:toy1}
        \end{subfigure}%
        ~ 
        \begin{subfigure}[b]{0.22\textwidth}
        \includegraphics[width=\textwidth]{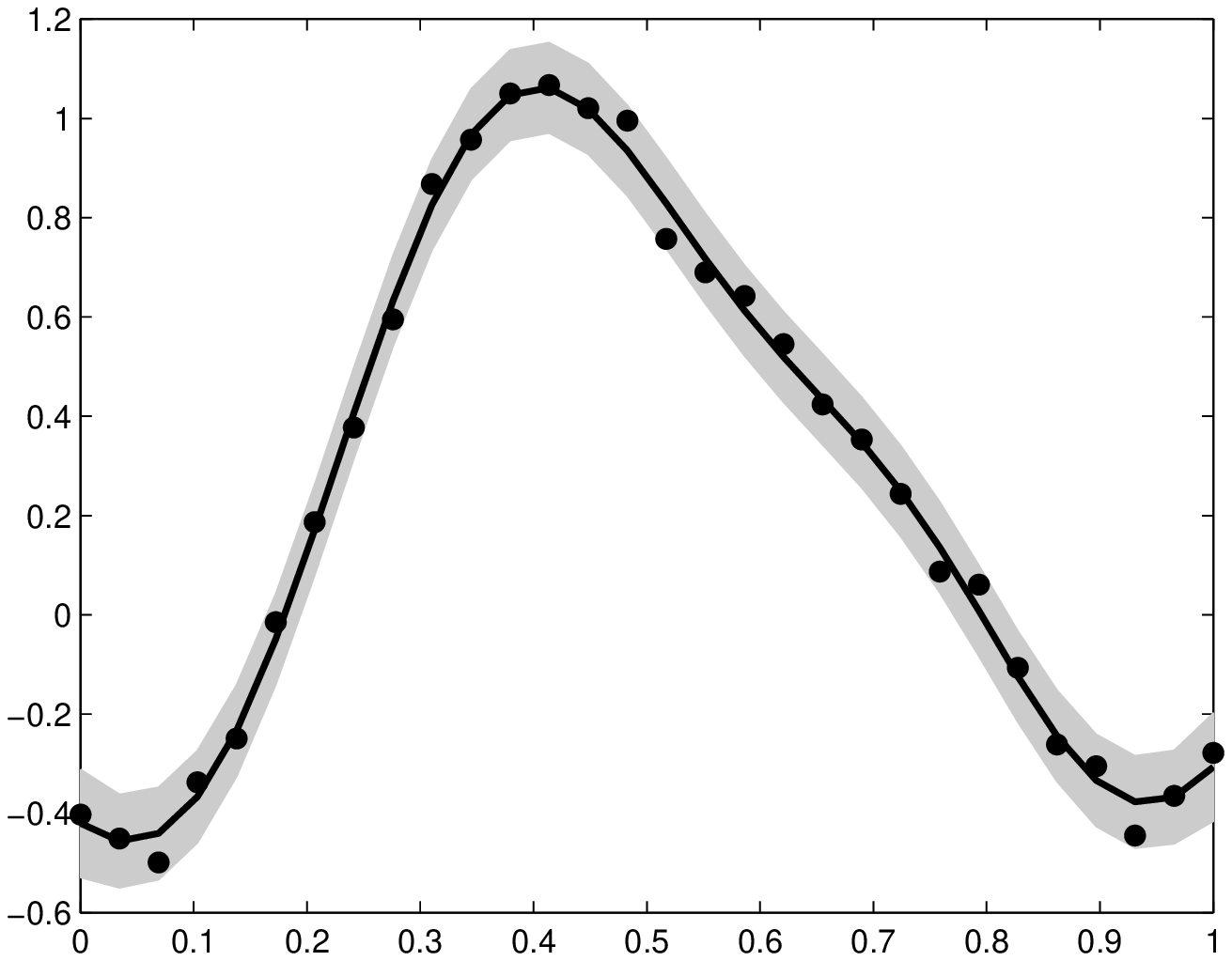}
   	    \caption{Ouput 1 toy 1}
        \label{toy1:output1}
        \end{subfigure}
        \begin{subfigure}[b]{0.22\textwidth}
        \includegraphics[width=\textwidth]{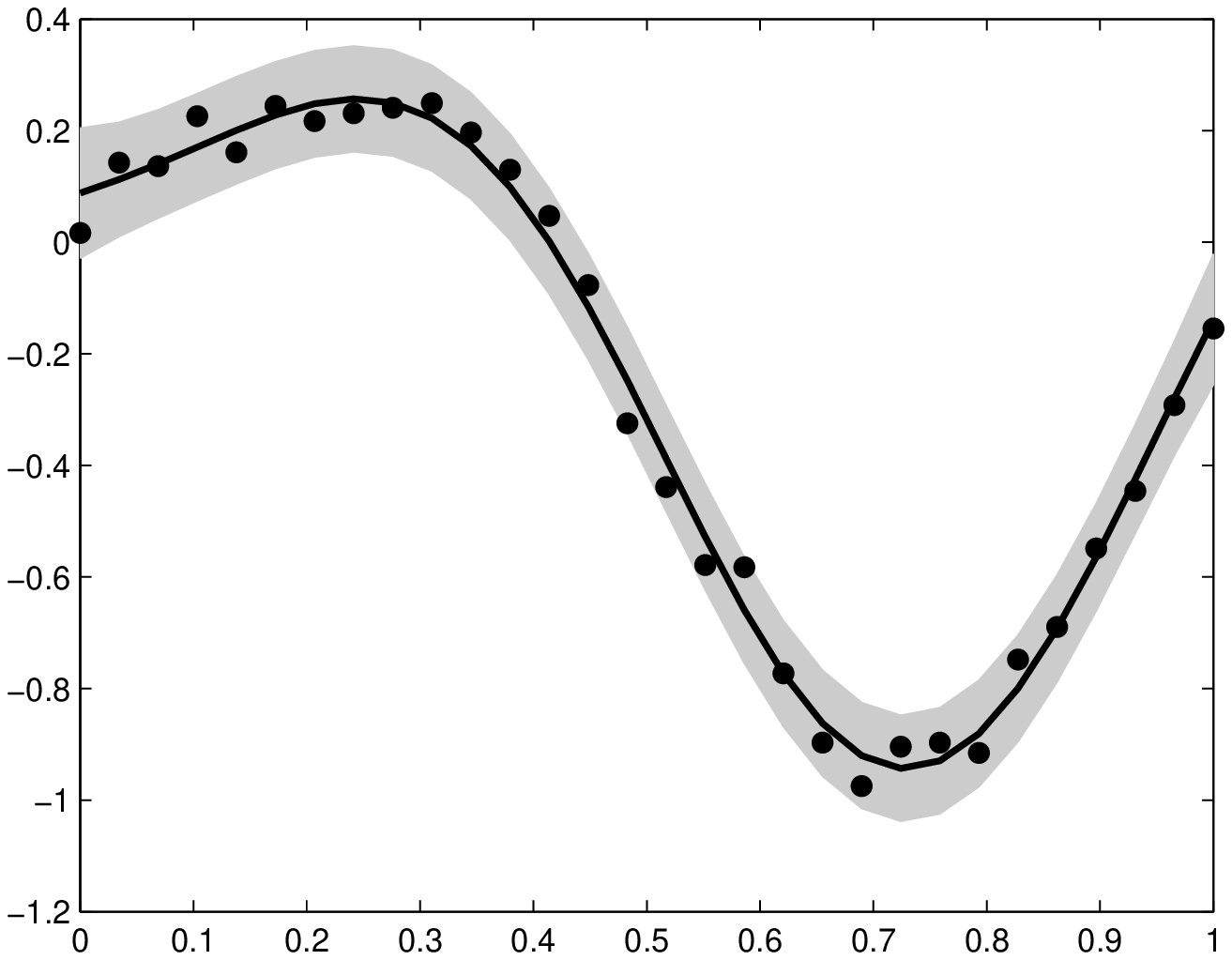}
        \caption{Ouput 2 toy 1}
        \label{toy1:output2}
        \end{subfigure}
        \begin{subfigure}[b]{0.22\textwidth}
        \includegraphics[width=\textwidth]{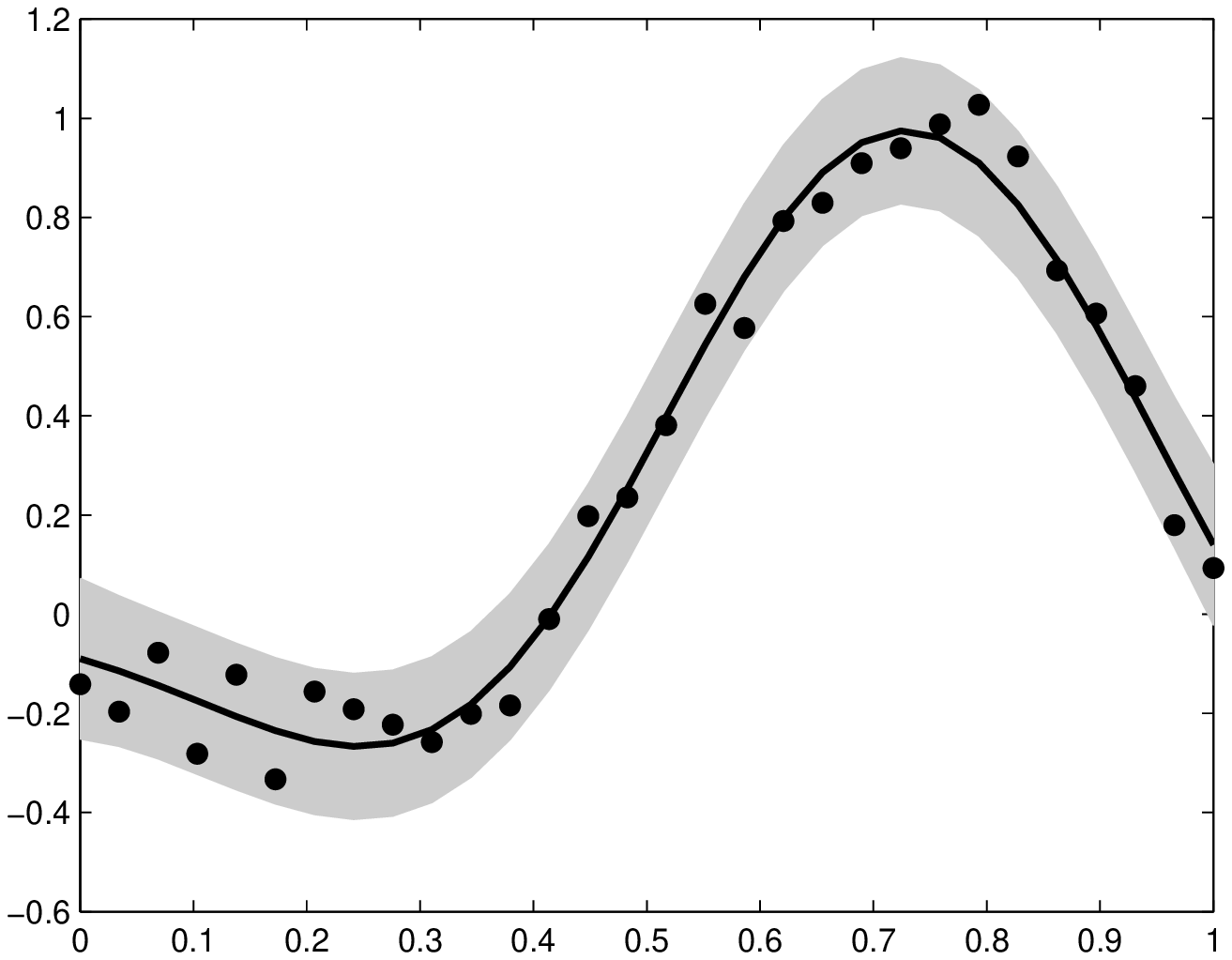}
        \caption{Ouput 3 toy 1}
        \label{toy1:output3}
        \end{subfigure}
        \caption{Results for model selection for example 1. Hinton diagram for $\mathbb{E}[Z_{d,q}]$ and, mean and two standard deviations for the predictions over the three outputs.}\label{fig:toygg}
\end{figure}

Figure \ref{fig:toygg} shows the results of model selection for this
experiment. We use a Hinton diagram to display the estimated value for
$\mathbb{E}[\boldZ]$, in Figure \ref{Hinton:toy1}. We notice from the
Hinton diagram that there are two main latent functions which are used
by the model to explain the data. The first column of the Hinton diagram corresponds to the second column of matrix $\boldZ$, while the second column of Hinton diagram corresponds to the first column of matrix $\boldZ$. The posterior mean functions for each output closely approximate the data, as shown in Figures \ref{toy1:output1} to \ref{toy1:output3}.

\paragraph{Example 2:} The second experiment is conducted using an ODE2 covariance function \citep{Alvarez:lfm09}. We generate data using $D = 3$, $Q=2$ and $\alpha =1 $. For each differential equation, we have the following values for the springs: $B_1= 4$, $B_2 = 1$, and $B_3 = 1$. The values for the dampers are $C_1= 0.5$, $C_2 = 4$, and $C_3 = 1$. Matrices $\boldZ$, and $\boldS$ are set to the following values

\[ \boldZ = \left[ \begin{array}{cc} 1 & 0\\ 0 & 1\\ 1 & 0 \end{array} \right] \quad \boldS = \left[ \begin{array}{cc} -2.61 & 0\\ 0 & 2.66\\ 1.10 & 0 \end{array} \right]
\]

The length-scales for the covariance functions of the latent Gaussian processes were set to $l_{1,1}=0.2$, and $l_{2,1} = 0.4$. We sample from the model, and generate 50 data points per output evenly spaced across the interval [0,5]. We truncate the number of latent functions to $Q=4$.

\begin{figure}[h]
        \centering
        \begin{subfigure}[b]{0.22\textwidth}
                \includegraphics[width=\textwidth]{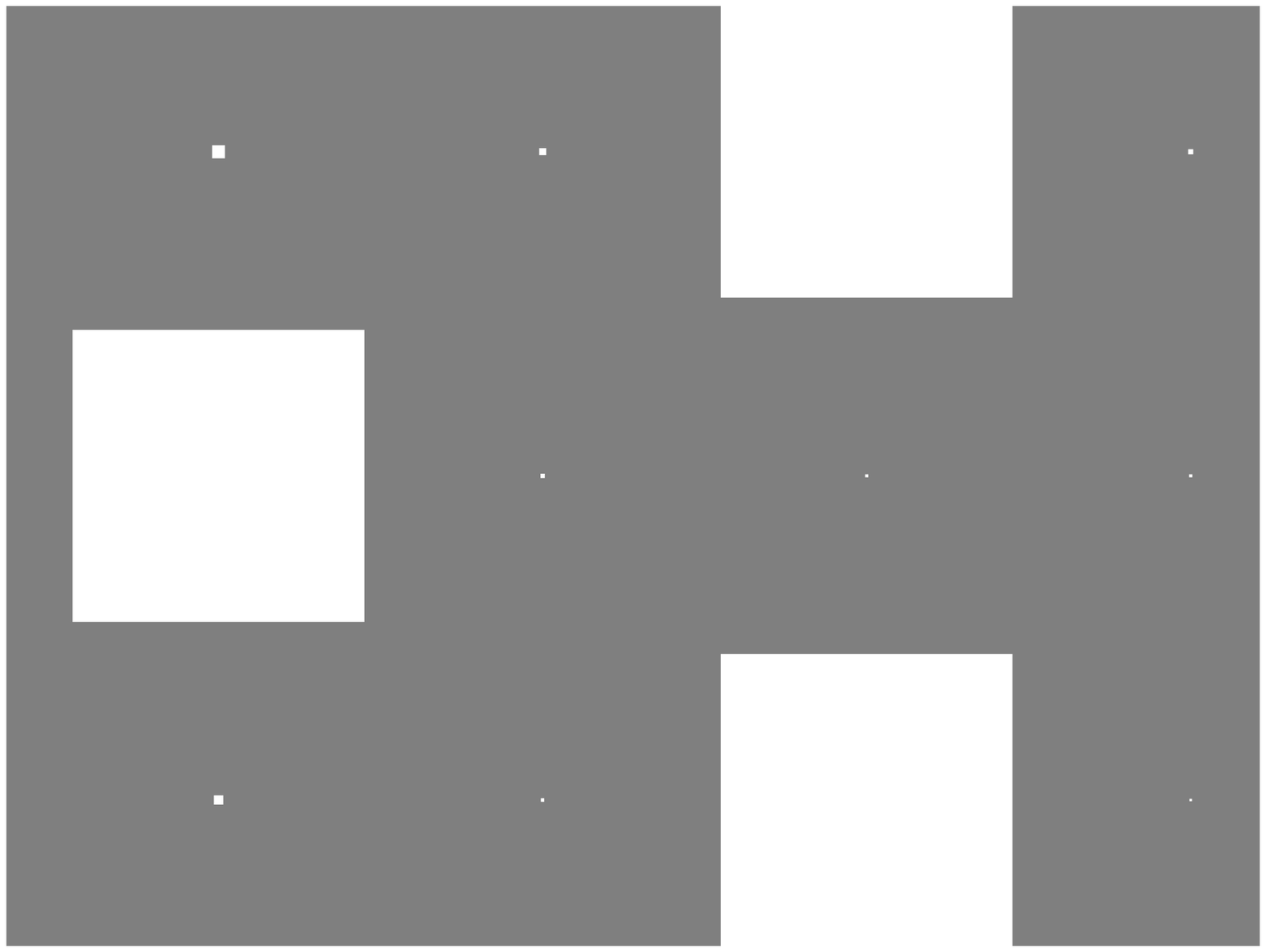}
                \caption{Hinton diagram toy 2}\label{hintontoy2}
        \end{subfigure}%
        ~ 
        \begin{subfigure}[b]{0.22\textwidth}
        \includegraphics[width=\textwidth]{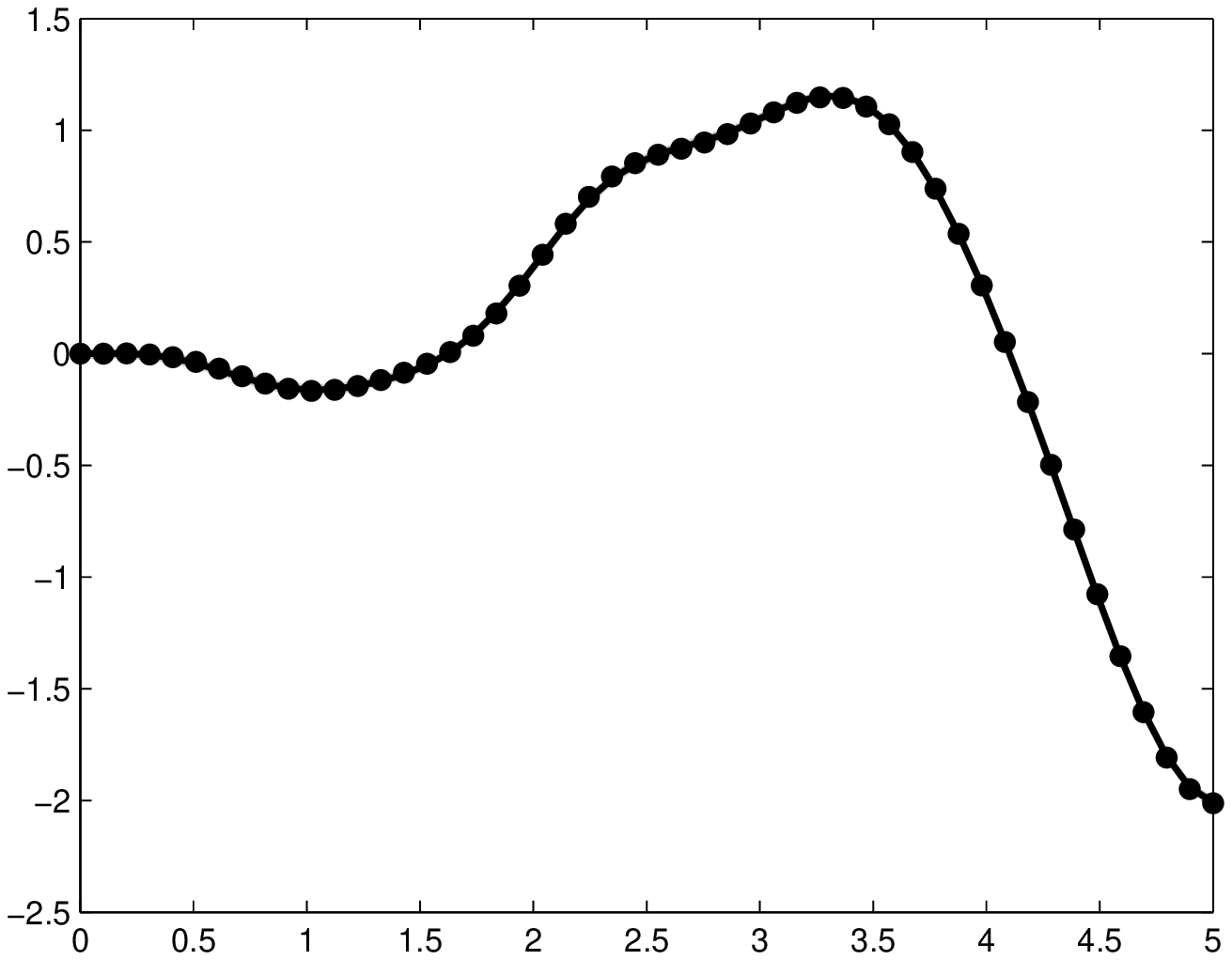}
        \caption{Output 1 toy 2}
        \label{toy2:output1}
        \end{subfigure}
        \begin{subfigure}[b]{0.22\textwidth}
           \includegraphics[width=\textwidth]{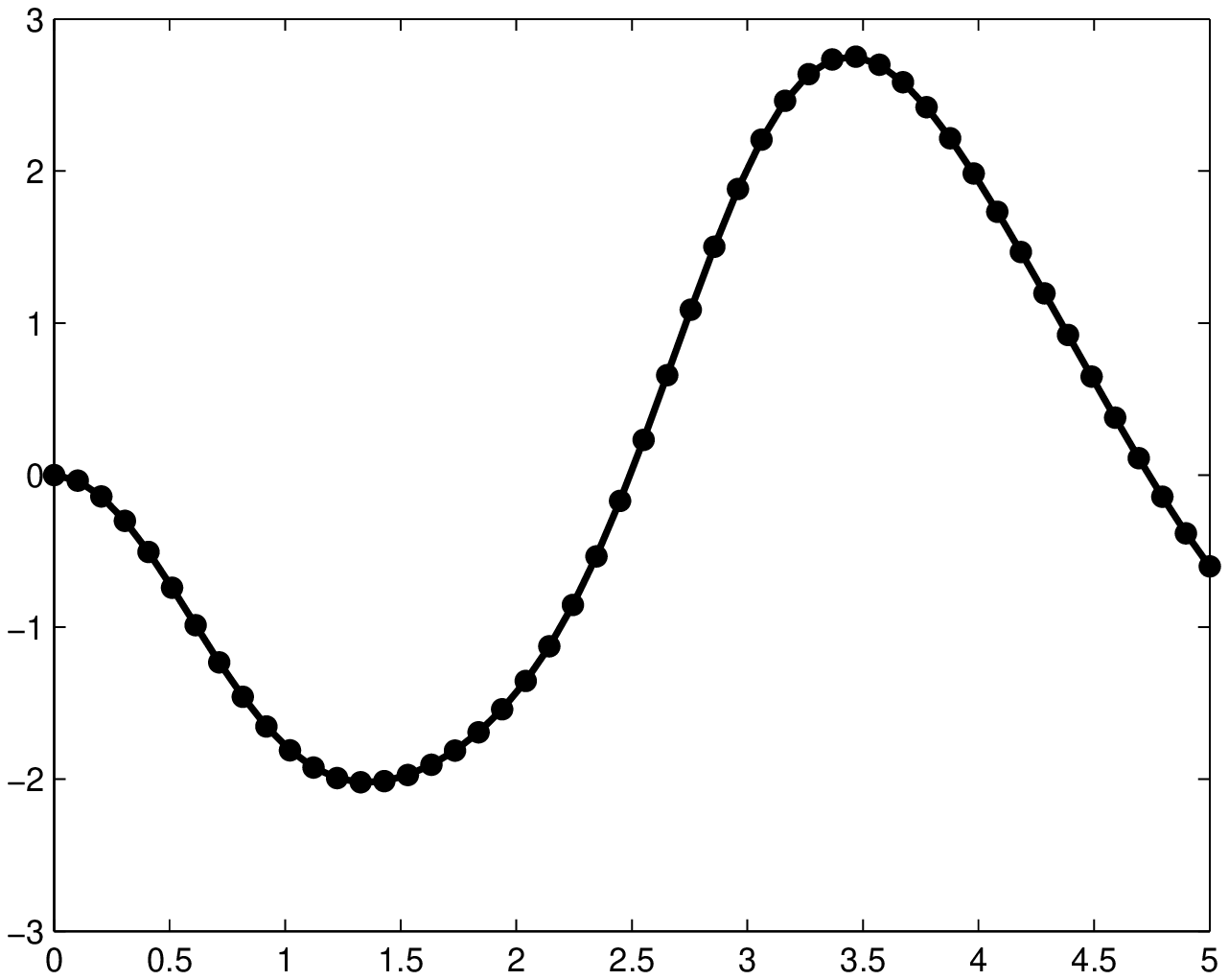}
           \caption{Output 2 toy 2}
           \label{toy2:output2}
        \end{subfigure}
        \begin{subfigure}[b]{0.22\textwidth}
           \includegraphics[width=\textwidth]{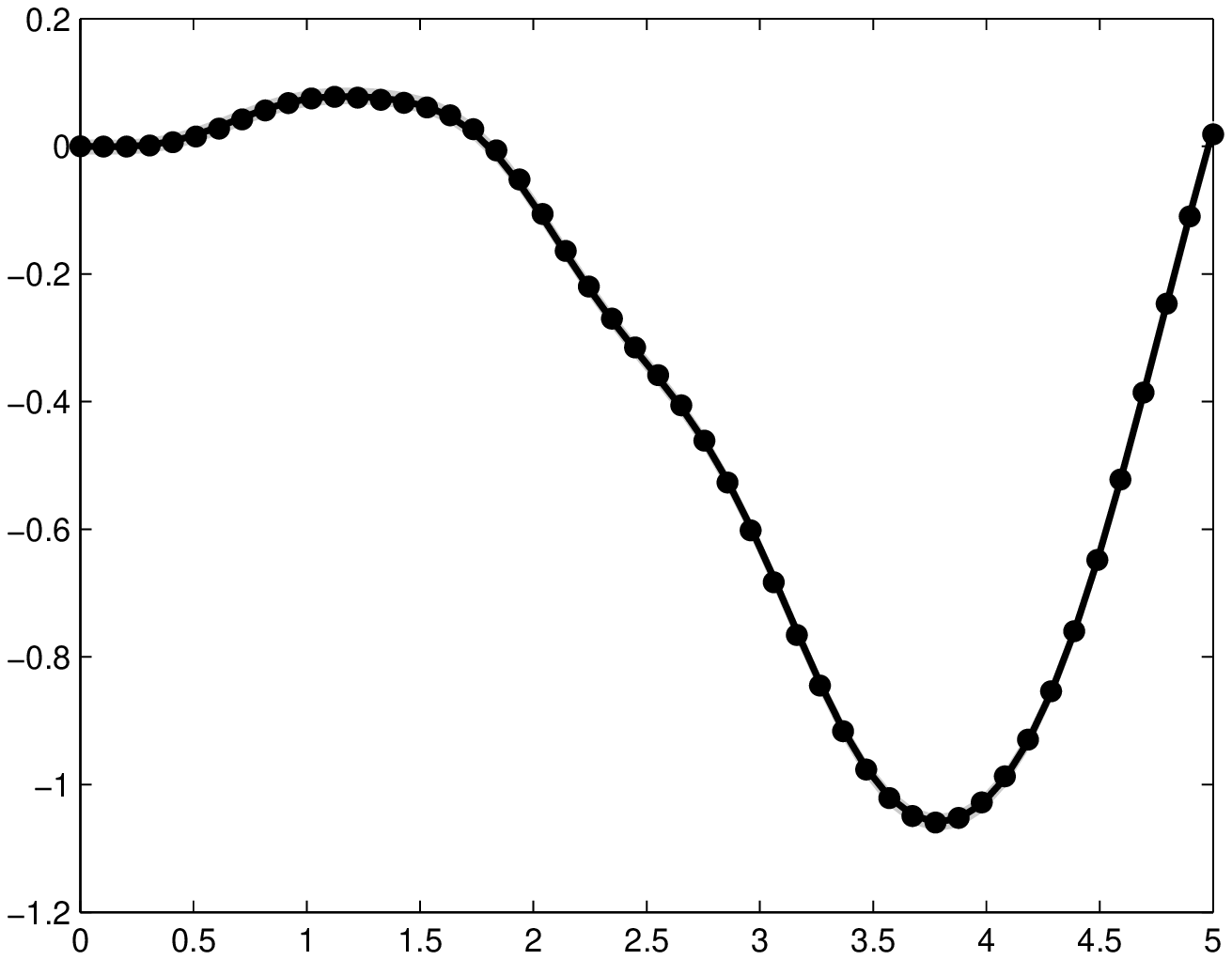}
           \caption{Output 3 toy 2}
           \label{toy2:output3}
        \end{subfigure}
        \caption{Results for model selection in toy example 2. In
          Figure \ref{hintontoy2}, Hinton diagram for
          $\mathbb{E}[Z_{d,q}]$. In Figures \ref{toy2:output1}
          to \ref{toy2:output3}, mean and two standard deviations for
          the predictive distribution over the three outputs.}\label{fig:toylfm}
\end{figure}

We perform the same evaluation as the one performed in example 1. The
Hinton diagram in Figure \ref{hintontoy2}, shows the values for
$\mathbb{E}[Z_{d,q}]$. We recover the structure imposed over the
original matrix $\boldZ$: columns first and third in the Hinton
diagram recover the ones and zeros in $\boldZ$, whereas columns second
and fourth have entries with very small values.

For this experiment, we used $y_d(t) = f_d(t)$ (we did not use an
independent process $w_d(t)$). The mean predictive function together
with the actual data is shown in Figures \ref{toy2:output1} to
\ref{toy2:output3}.

In the following sections, we evaluate the performance of the proposed
model selection method in human motion capture data and gene
expression data.

\subsection{Human motion capture data}
In this section, we evaluate the performance of the proposed method
compared to the Deterministic Training Conditional Variational
(DTCVAR) inference procedure proposed in \citet{Alvarez:VIKs:report:2009}. DTCVAR also uses inducing variables
for reducing computational complexity within a variational framework,
but assumes full connectivity between the latent functions and the
output functions (meaning that $Z_{d,q}=1$, for all $q$, and
$d$). Parameters for all the kernel functions employed are learned
using scaled conjugate gradient optimization. 

\subsubsection{Performance of the model in terms of changing number of outputs and the truncation level}
\begin{figure}[h]
        \centering
        \begin{subfigure}[b]{0.22\textwidth}
                \includegraphics[width=\textwidth, height=4cm]{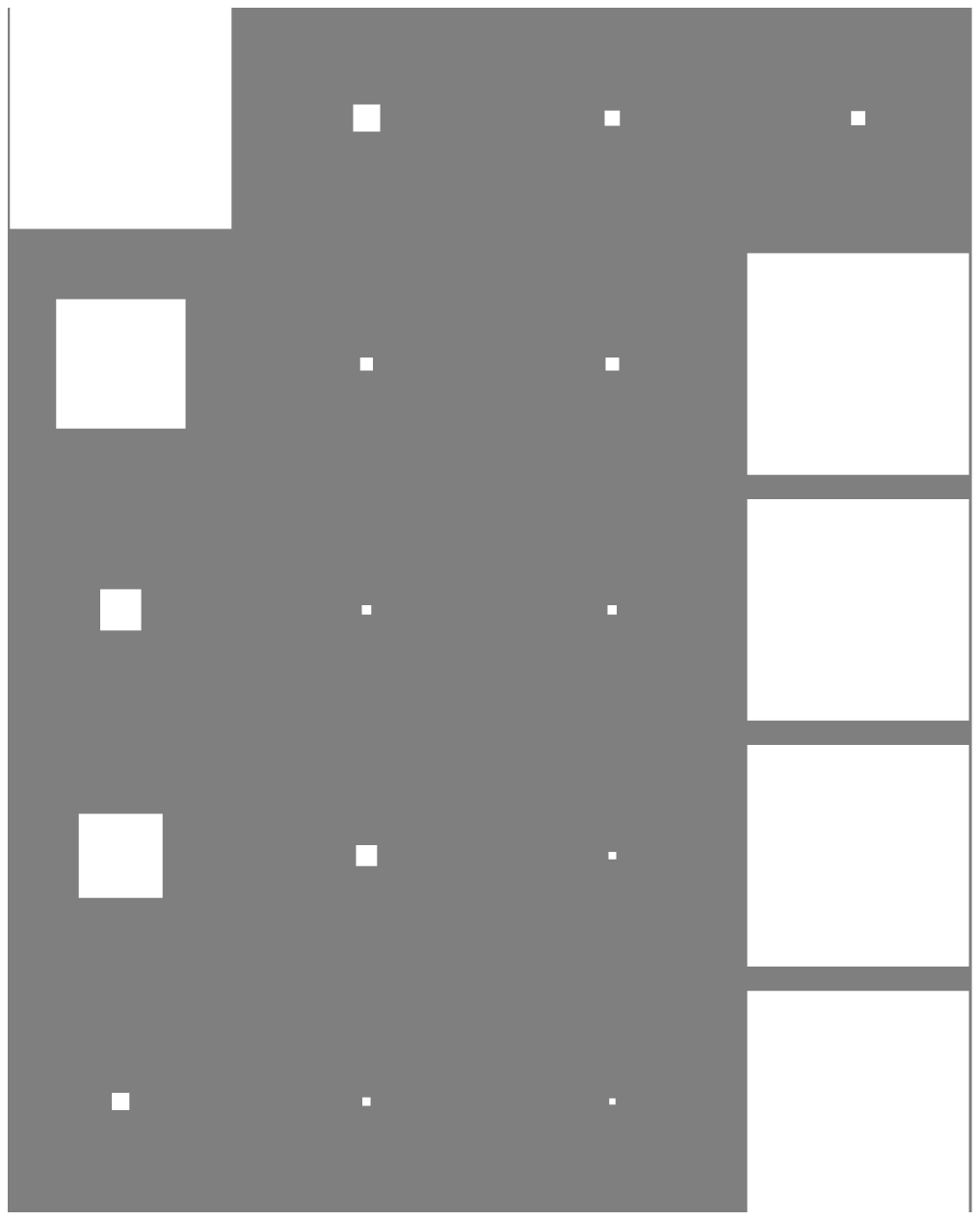}
                \caption{D=5, Q=4}
        \end{subfigure}%
        ~ 
        \begin{subfigure}[b]{0.22\textwidth}
        \includegraphics[width=\textwidth,height=4cm]{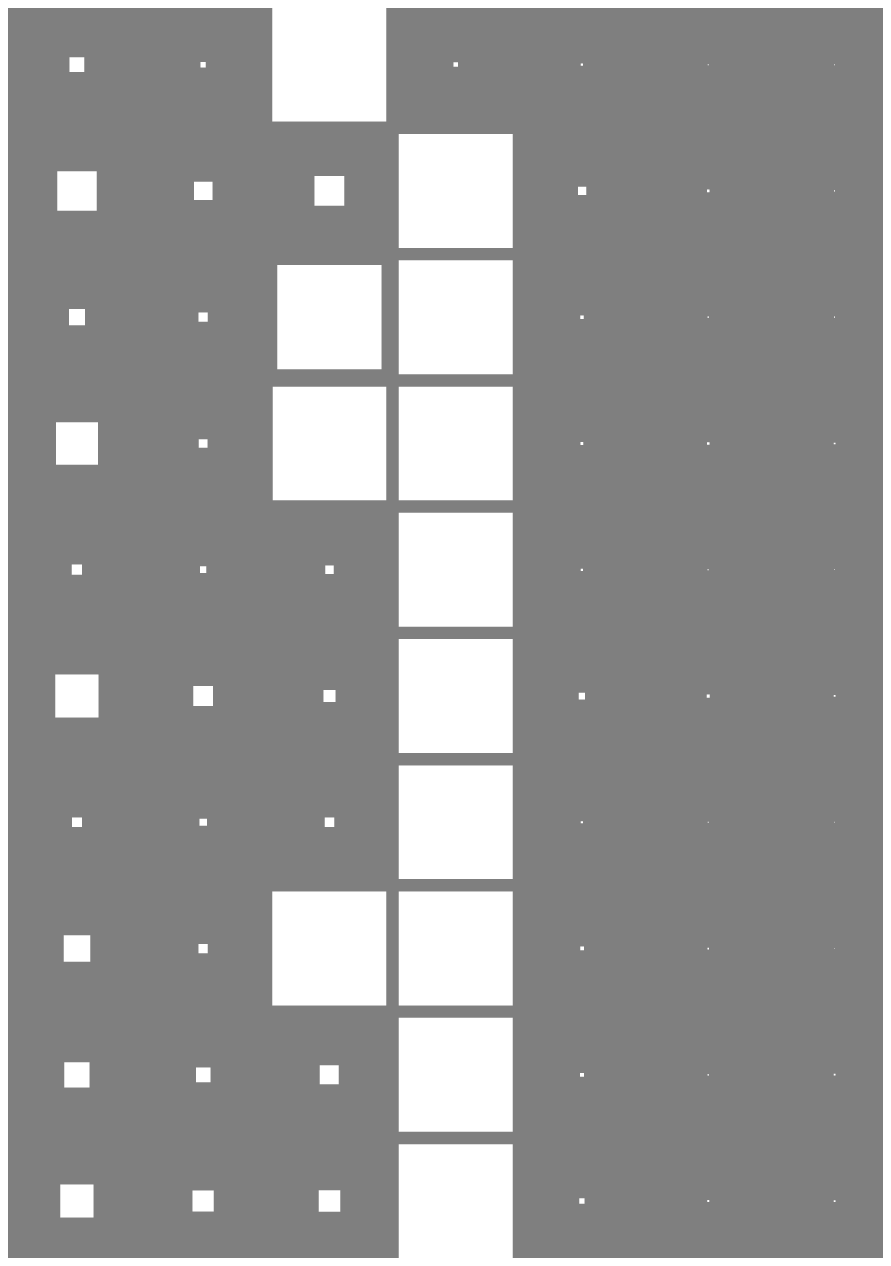}
        \caption{D=10, Q=7}
        \end{subfigure}
        \begin{subfigure}[b]{0.22\textwidth}
           \includegraphics[width=\textwidth,height=4cm]{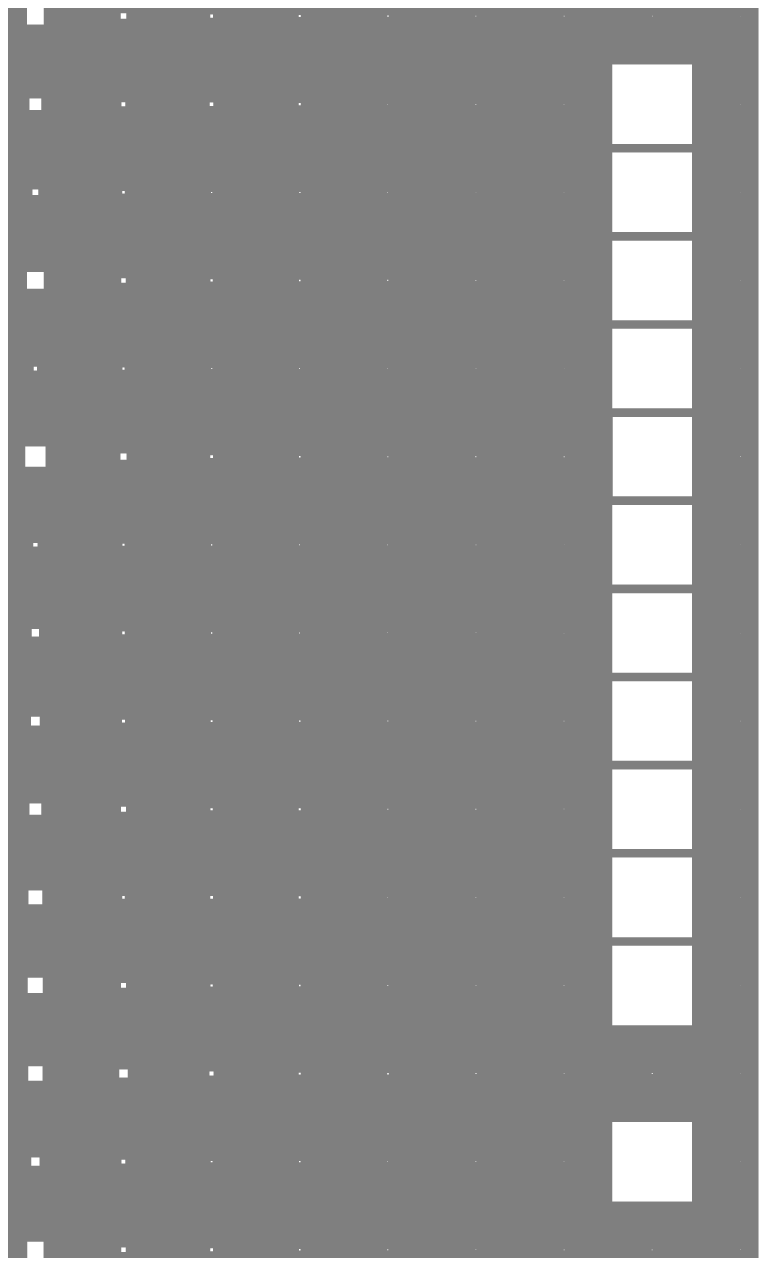}
           \caption{D=15, Q=9}
        \end{subfigure}
        \begin{subfigure}[b]{0.22\textwidth}
           \includegraphics[width=\textwidth,height=4cm]{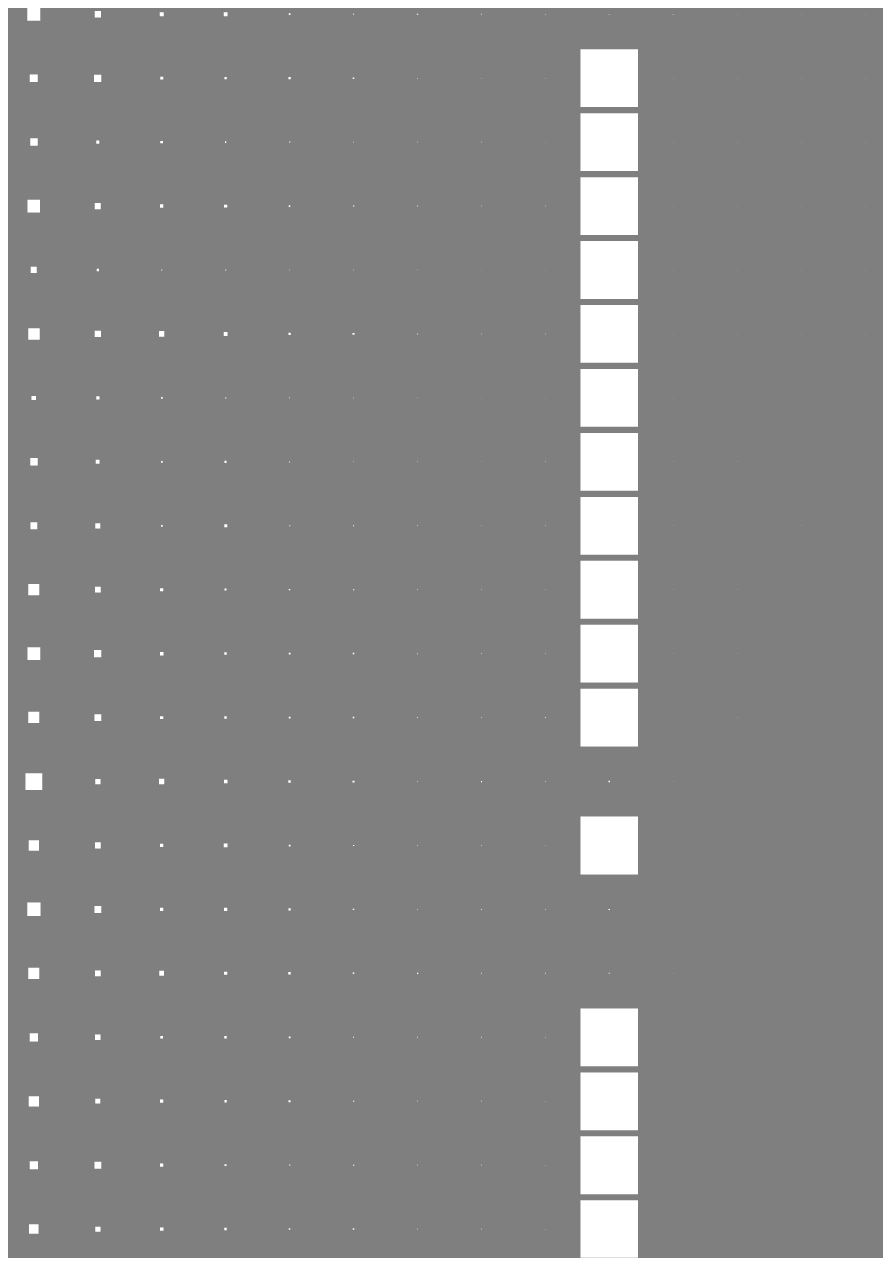}
           \caption{D=20, Q=14}
        \end{subfigure}
        \caption{Hinton diagrams of $\mathbb{E}[\boldZ]$ for each pair of outputs and latent functions tested in table \ref{tab:PerIncInOut}.}\label{fig:hintonwal}
\end{figure}

We use the Carnegie Mellon University's Graphics Lab motion-capture
motion capture database available at http://mocap.cs.cmu.edu.
Specifically, we consider the movement walking from subject 35 (motion
01). From this movement, we select 20 channels from the 62 available
(we avoid channels where the signals were just noise or a straight
line). Then, we take 45 frames for training and the rest 313 frames
are left out for testing. Table \ref{tab:PerIncInOut} shows a
performance comparison between our proposed model and a model trained
using DTCVAR, taking into account different types of covariance
functions. Performance is measured using standardized mean square
error (SMSE) and mean standardized log loss (MSLL) over the test set
for different combinations of number of outputs ($D$) and number of
latent functions ($Q$). Similar results are obtained by both models
using the GS covariance function. Even-though, the model based on DTCVAR
outperforms the proposed one in the cases $D=10,Q=7$, and
$D=15,Q=9$, the DTCVAR approach uses all latent functions, while the
IBP approach uses only two latent functions, as shown in Figure
\ref{fig:hintonwal}.
\begin{table*}[h]
\centering
\begin{tabular}{cccccc}
\hline
Inference setup & Measure & $D$=5, $Q$=4 & $D$=10, $Q$=7 & $D$=15, $Q$=9 & $D$=20, $Q$=14\\
\hline
GS & SMSE & 0.241 & 0.1186 & 0.3072 & 0.3796\\
& MSLL & -1.1334 & -1.6924 & -1.0322 & -0.9196 \\
\hline
IBP + GS & SMSE & 0.1538 & 0.3267 & 0.3494 & 0.3605\\
& MSLL & -1.6083 & -1.0140 & -0.8819 & -0.8118\\
\hline
\end{tabular}
\caption{Standardized mean square error (SMSE) and mean standardized log loss (MSLL) for different number of outputs and latent functions.}
\label{tab:PerIncInOut}
\end{table*}

For the other two cases ($D=5,Q=4$, and $D=20, Q=14$), the proposed
method presents a similar performance compared to the model estimated
by DTCVAR using all latent functions, showing that to obtain an
adequate approximation of the output data we do not require to use the
maximum number of the latent functions.

\subsubsection{Performance for different kernel functions}
In this section, we compare the performance of the proposed model and
the one trained using DTCVAR with different covariance functions over
the same dataset. In this case, we consider the walking movement from
subject 02 motion 01. From the 62 channels, we select 15 for this
experiment. We assume a maximum of nine latent functions, and
make a comparison between the GS and the ODE2 covariance
functions. The latter is used because human motion data consists of
recordings of an skeleton's joint angles across time, which summarize
the motion. We can use a set of second order differential equations to
describe such motion.  Table \ref{tab:CompODEGG} reports the SMSE and
MSLL measures for each type of training method and covariance
function. Our proposed method presents better results, with the ODE2
kernel being the kernel that best explains the data.

\begin{figure}[H]
        \centering
        \begin{subfigure}[b]{0.22\textwidth}
                \includegraphics[width=\textwidth, height=6cm]{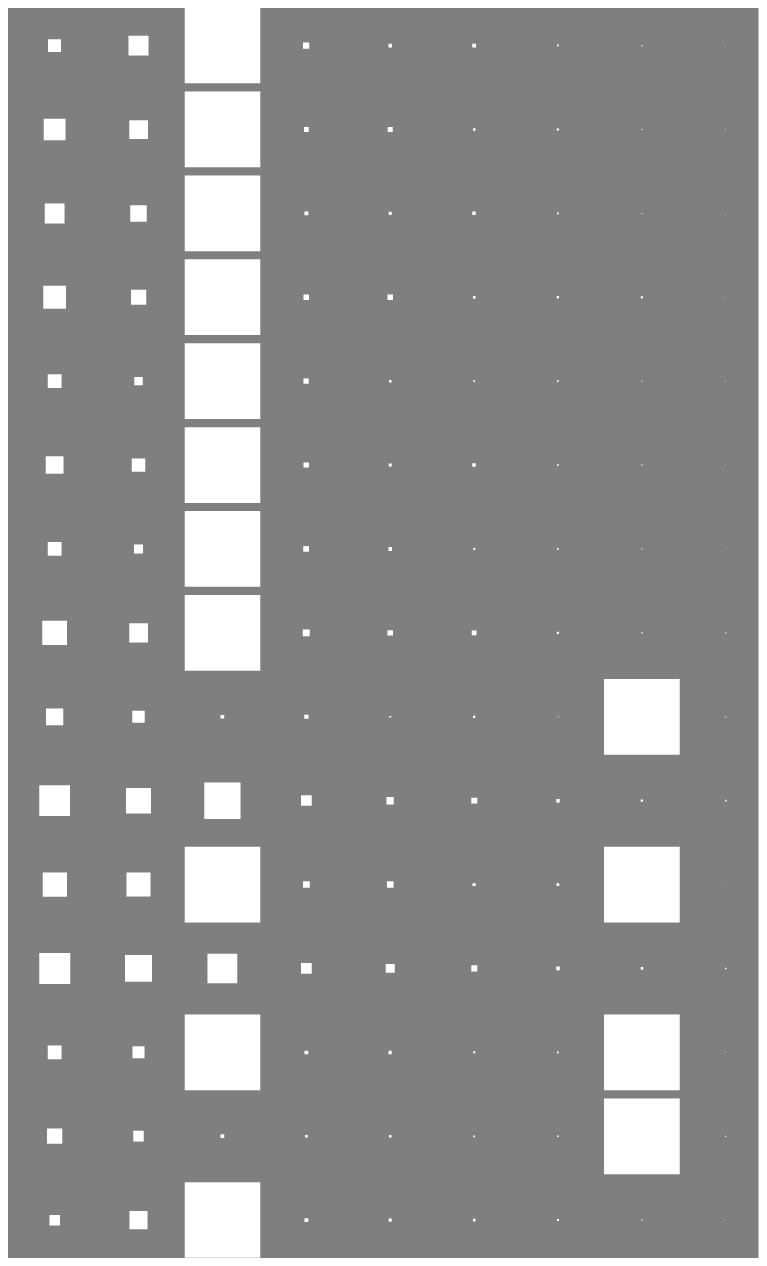}
                \caption{IBP + ODE2 }
                \label{fig:gull}
        \end{subfigure}%
        ~ 
        \begin{subfigure}[b]{0.22\textwidth}
                \includegraphics[width=\textwidth,height=6cm]{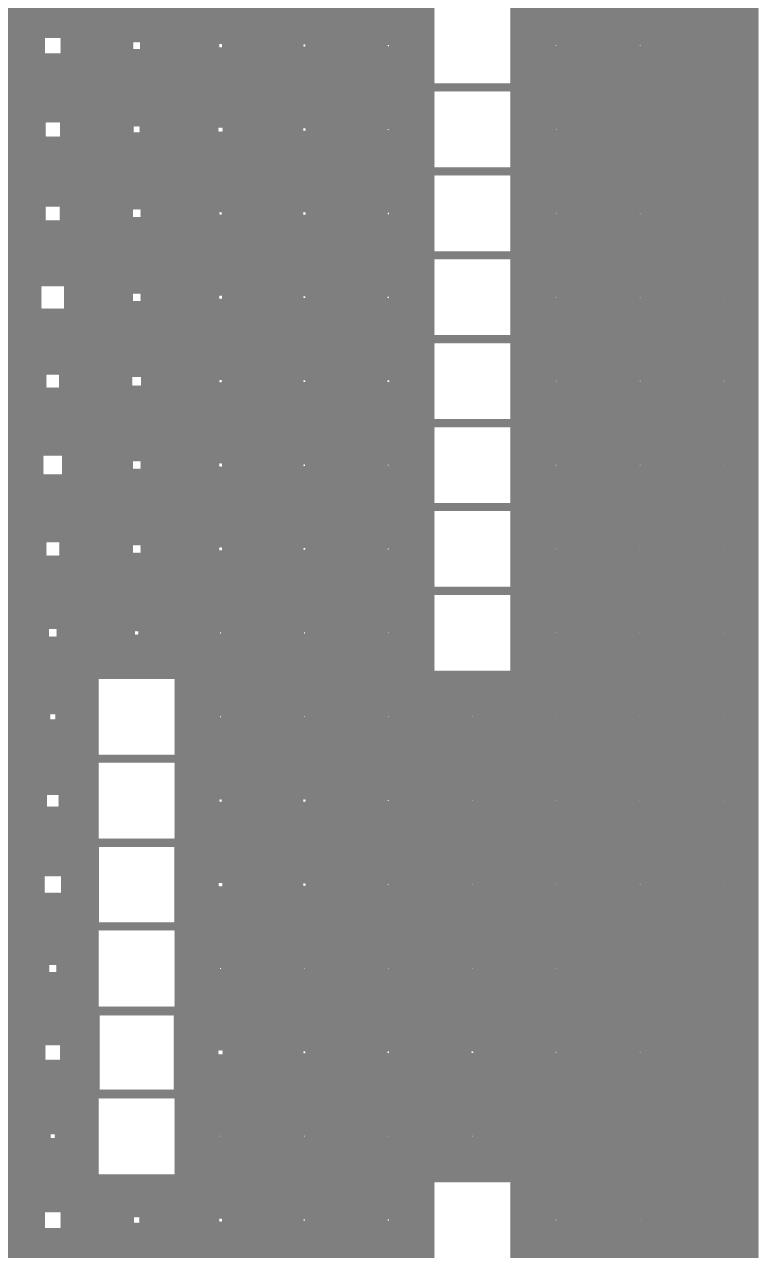}
                \caption{IBP + GS}
                \label{fig:tiger}
        \end{subfigure}
        \caption{Hinton diagrams for the IBP variational approximation using (a) ODE2 and (b) GS covariance functions.}\label{fig:hintoncomp}
\end{figure}

\begin{table}[h]
\centering
\begin{tabular}{ccccc}
\hline
& ODE2 & IBP + ODE2 & GS & IBP + GS\\
\hline
SMSE & 0.5463 & 0.2087 & 0.5418 & \bf{0.1790}\\
SMLL & -0.6547 & \bf{-1.2725} & -0.7863 & -1.1993\\
\hline
\end{tabular}
\caption{Standardized mean square error (SMSE) and mean standardized
  log loss (MSLL) for different models and different kernel functions.}
\label{tab:CompODEGG}
\end{table}
%
Comparing the Hinton diagrams from both covariance functions (see Figure \ref{fig:hintoncomp}), we find similar results. For example, there is a similar composition of elements between columns 3 and 8 from the Hinton diagram of the ODE2 kernel, with columns 6 and 2 from the Hinton diagram of the GS kernel. Both covariance functions try to unveil a similar interconnection between outputs and latent functions.

\begin{figure}[H]
        \centering
        \begin{subfigure}[b]{0.24\textwidth}
                \includegraphics[width=\textwidth]{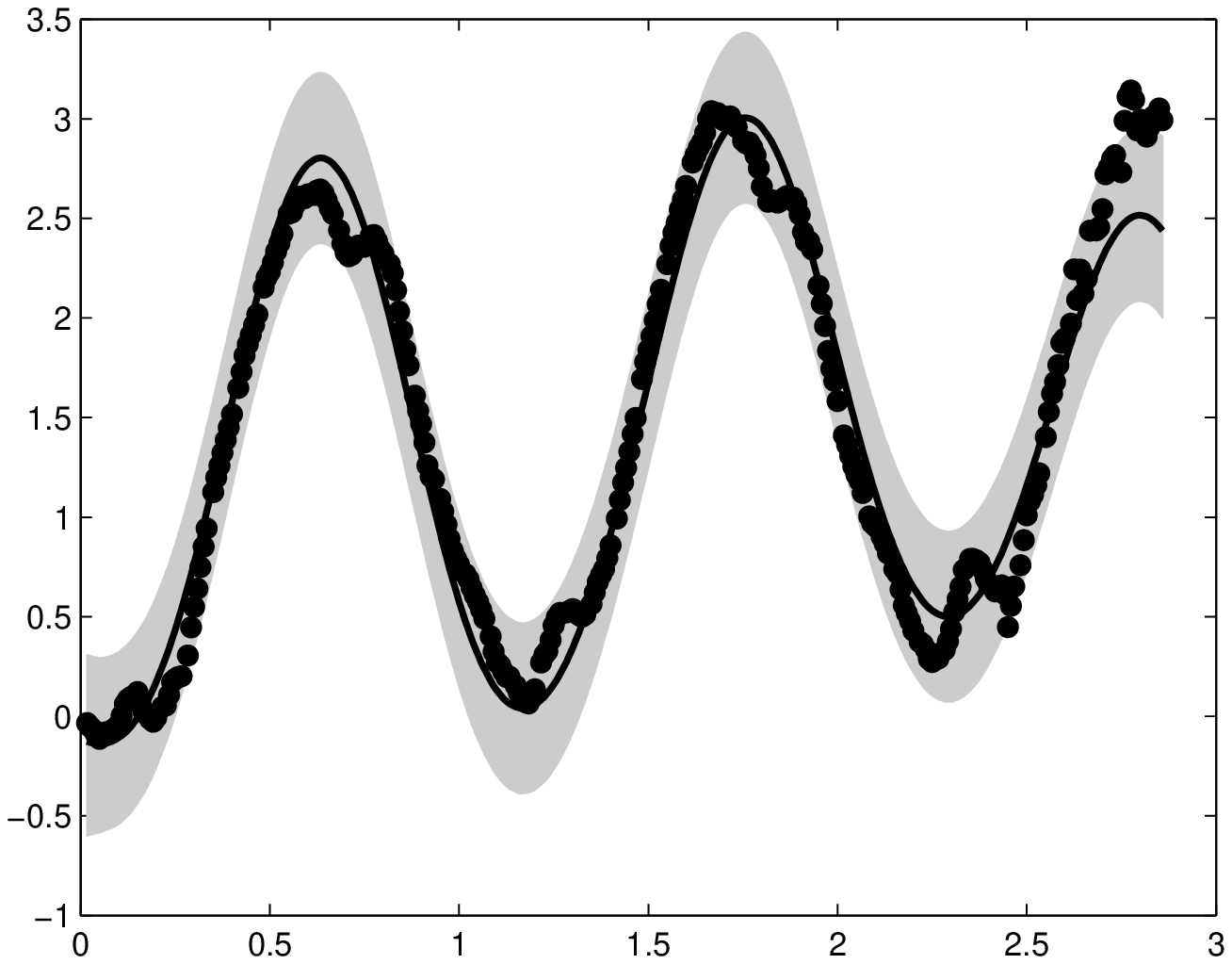} 
                \caption{Channel 5 Root}
        \end{subfigure}%
        ~ 
        \begin{subfigure}[b]{0.24\textwidth}
                \includegraphics[width=\textwidth]{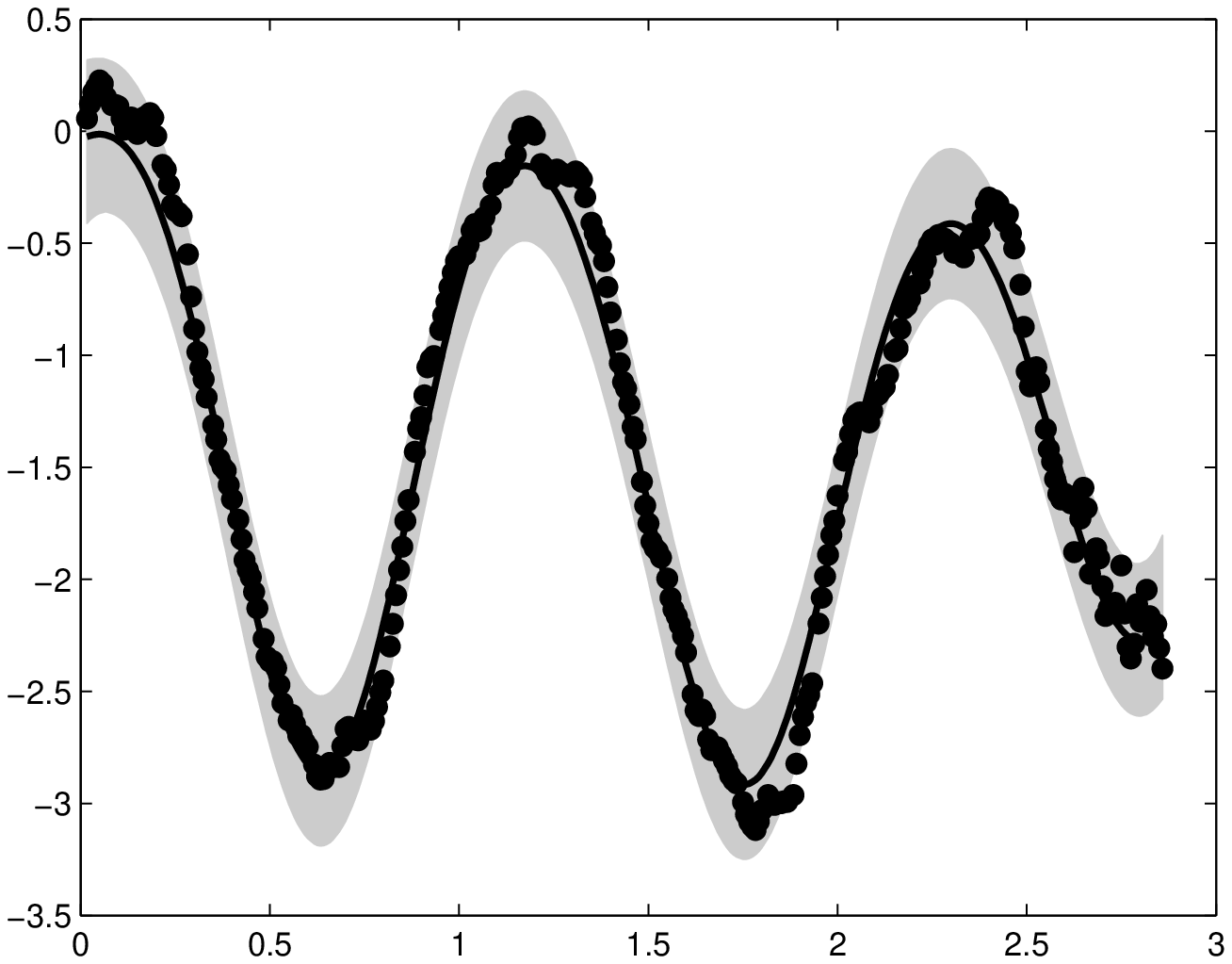} 
                \caption{Channel 2 Lower Back}
        \end{subfigure}
        \begin{subfigure}[b]{0.23\textwidth}
           \includegraphics[width=\textwidth]{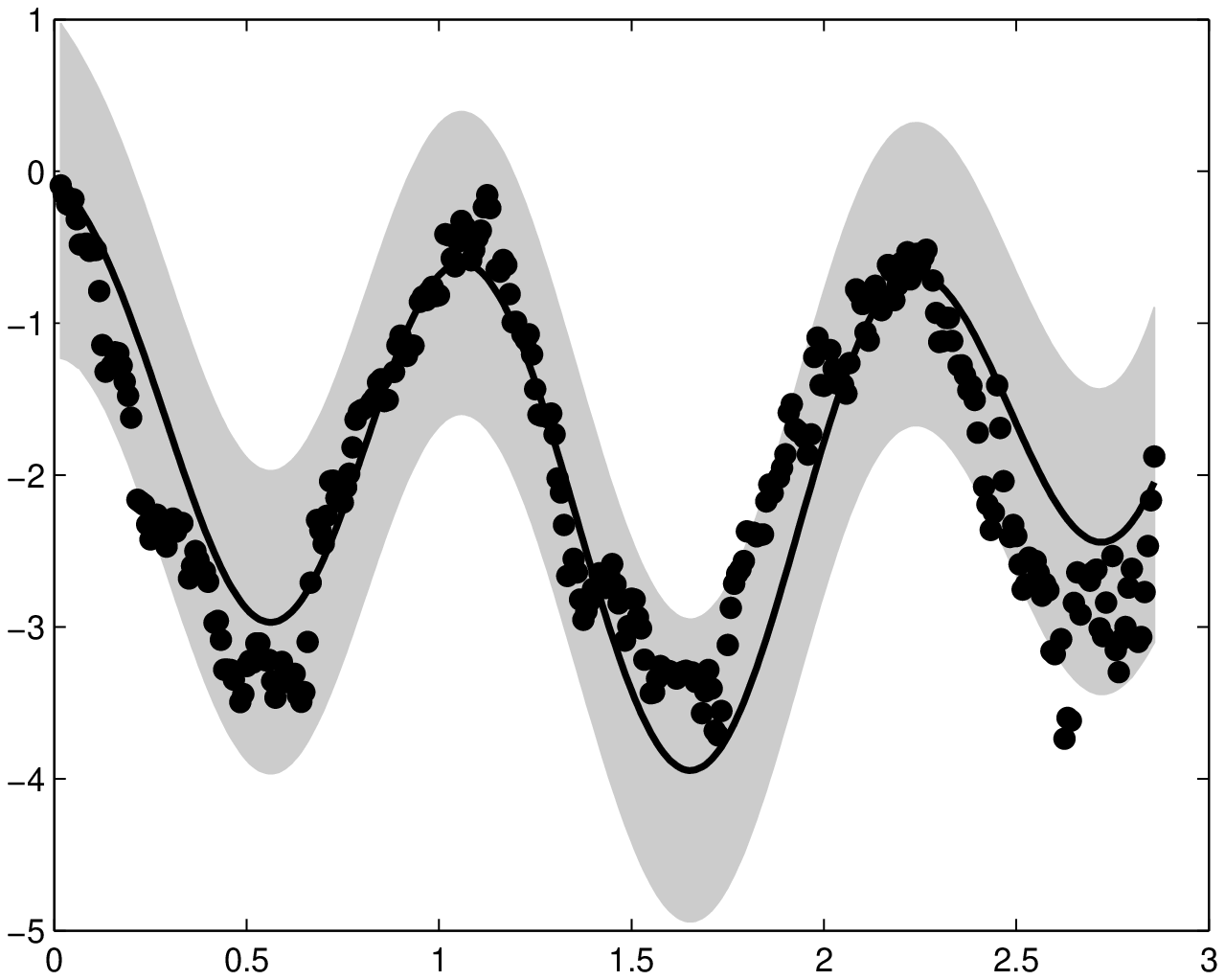} 
           \caption{Channel 3 Lower back}
        \end{subfigure}
        \begin{subfigure}[b]{0.23\textwidth}
           \includegraphics[width=\textwidth]{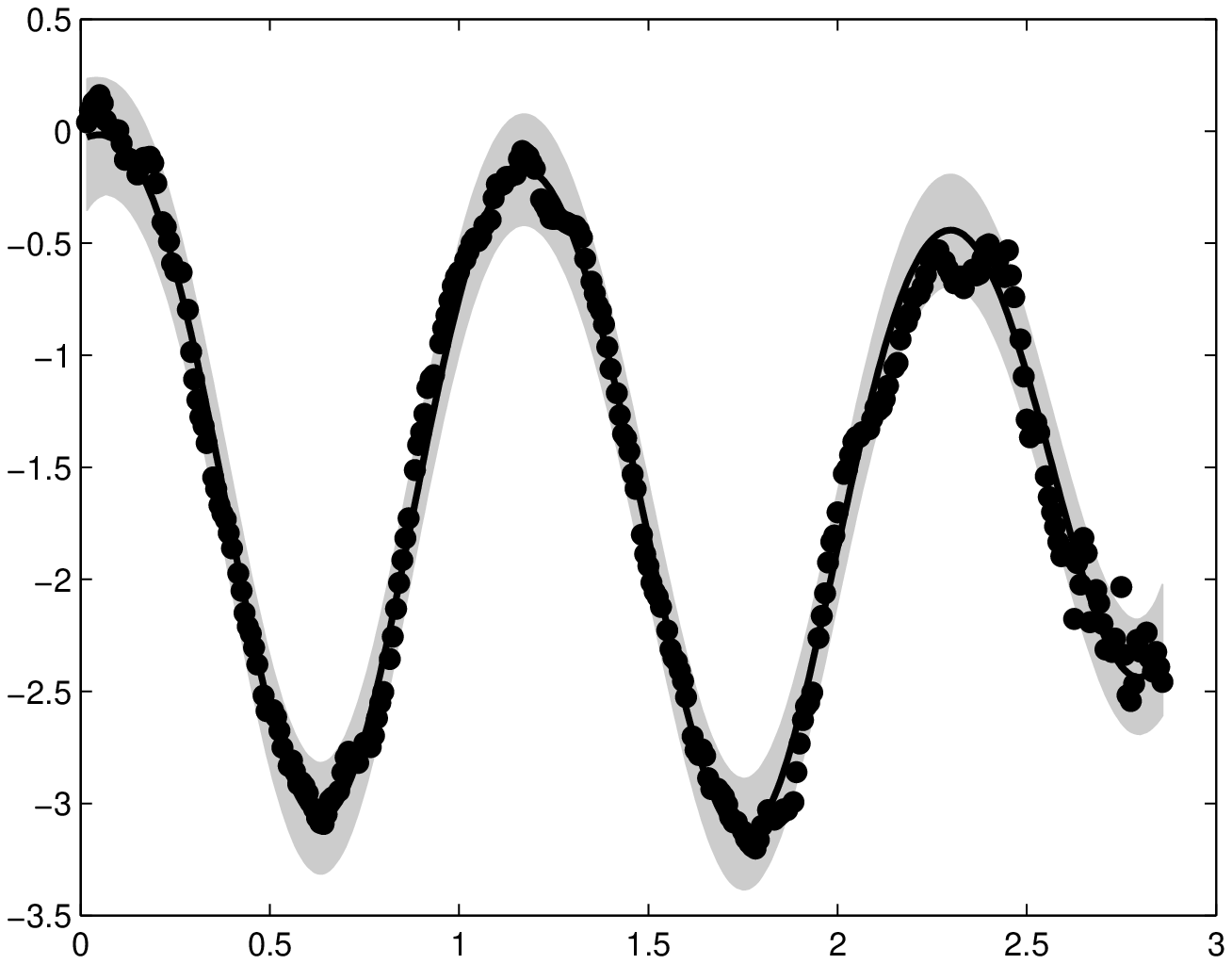} 
           \caption{Channel 2 Thorax}
        \end{subfigure}\\
        \begin{subfigure}[b]{0.24\textwidth}
                \includegraphics[width=\textwidth]{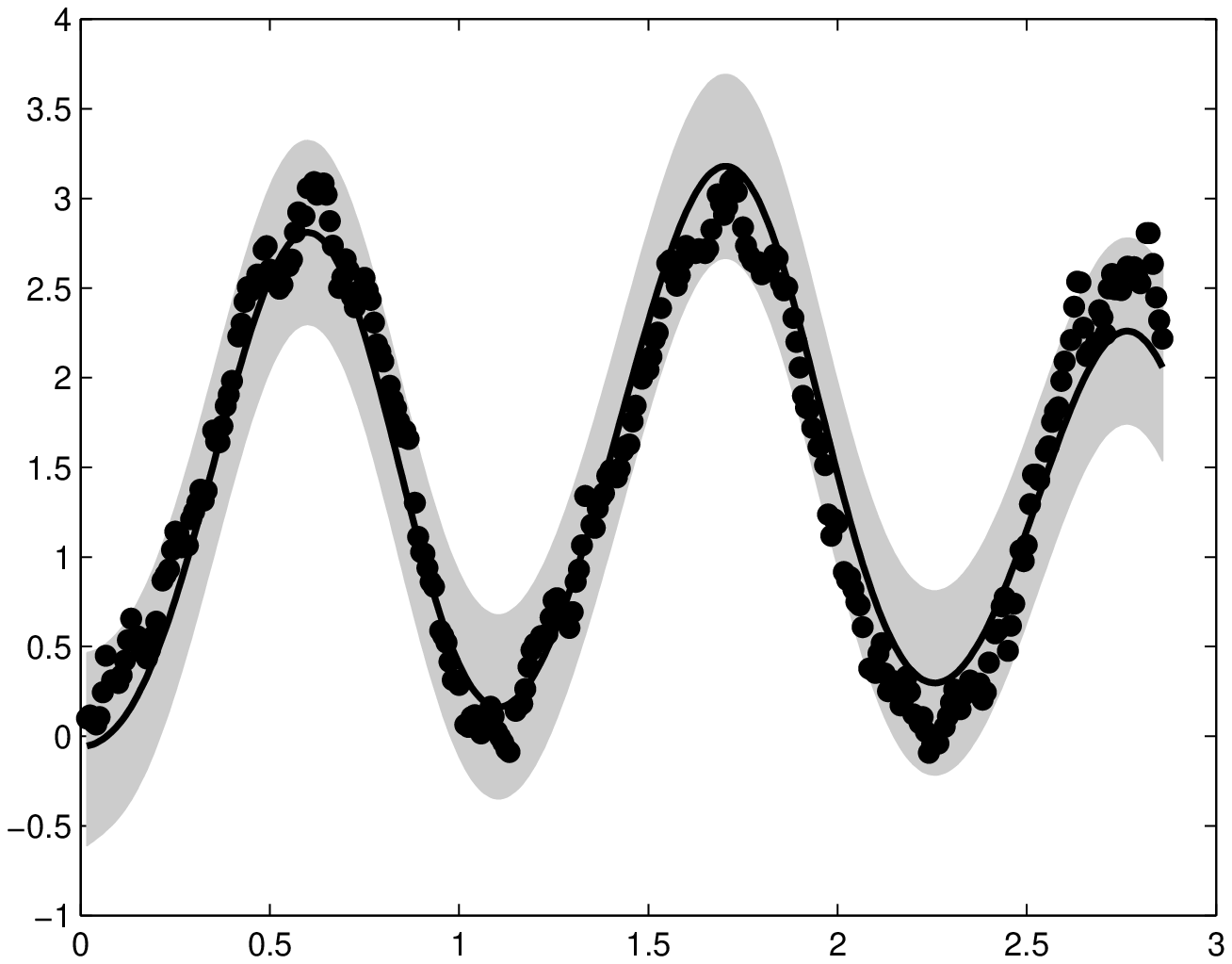} 
                \caption{Channel 3 Thorax}
        \end{subfigure}%
        ~ 
        \begin{subfigure}[b]{0.24\textwidth}
                \includegraphics[width=\textwidth]{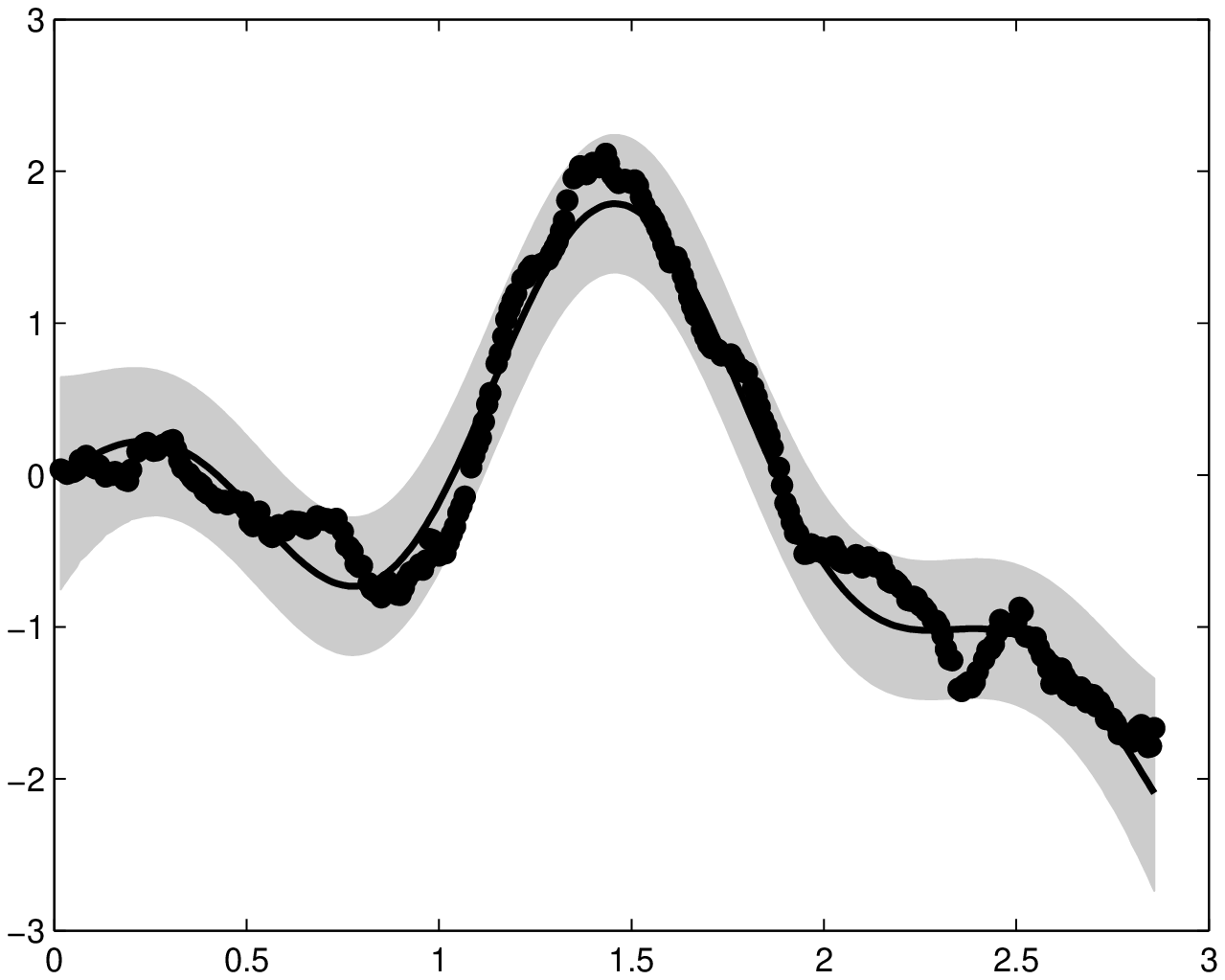} 
                \caption{Channel 2 Head}
        \end{subfigure}
        \caption{Mean (solid line) and two standard deviations (gray shade) for predictions over six selected outputs from IBP + ODE2 trained model.}\label{fig:outputIBPLFM}
\end{figure}

Figure \ref{fig:outputIBPLFM} shows the Gaussian process mean and
variance for the predictive distribution of six outputs from the model
inferred from IBP + ODE2. In most of the predictions, the model
explains the testing data points with adequate accuracy.

\begin{figure}[ht!]
        \centering
        \begin{subfigure}[b]{0.24\textwidth}
                \includegraphics[width=\textwidth]{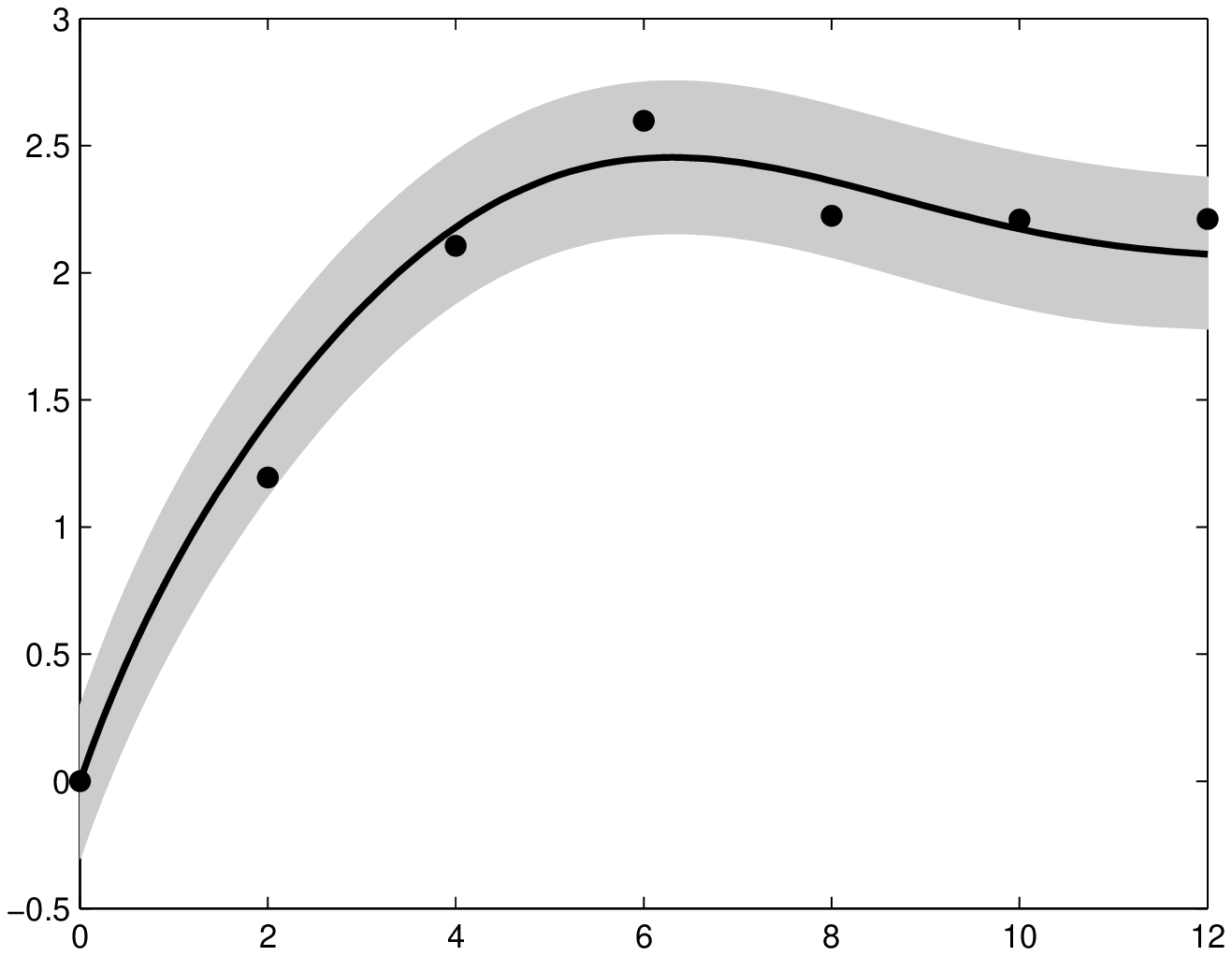} 
                \caption{DDB2}
        \end{subfigure}%
        ~ 
        \begin{subfigure}[b]{0.24\textwidth}
                \includegraphics[width=\textwidth]{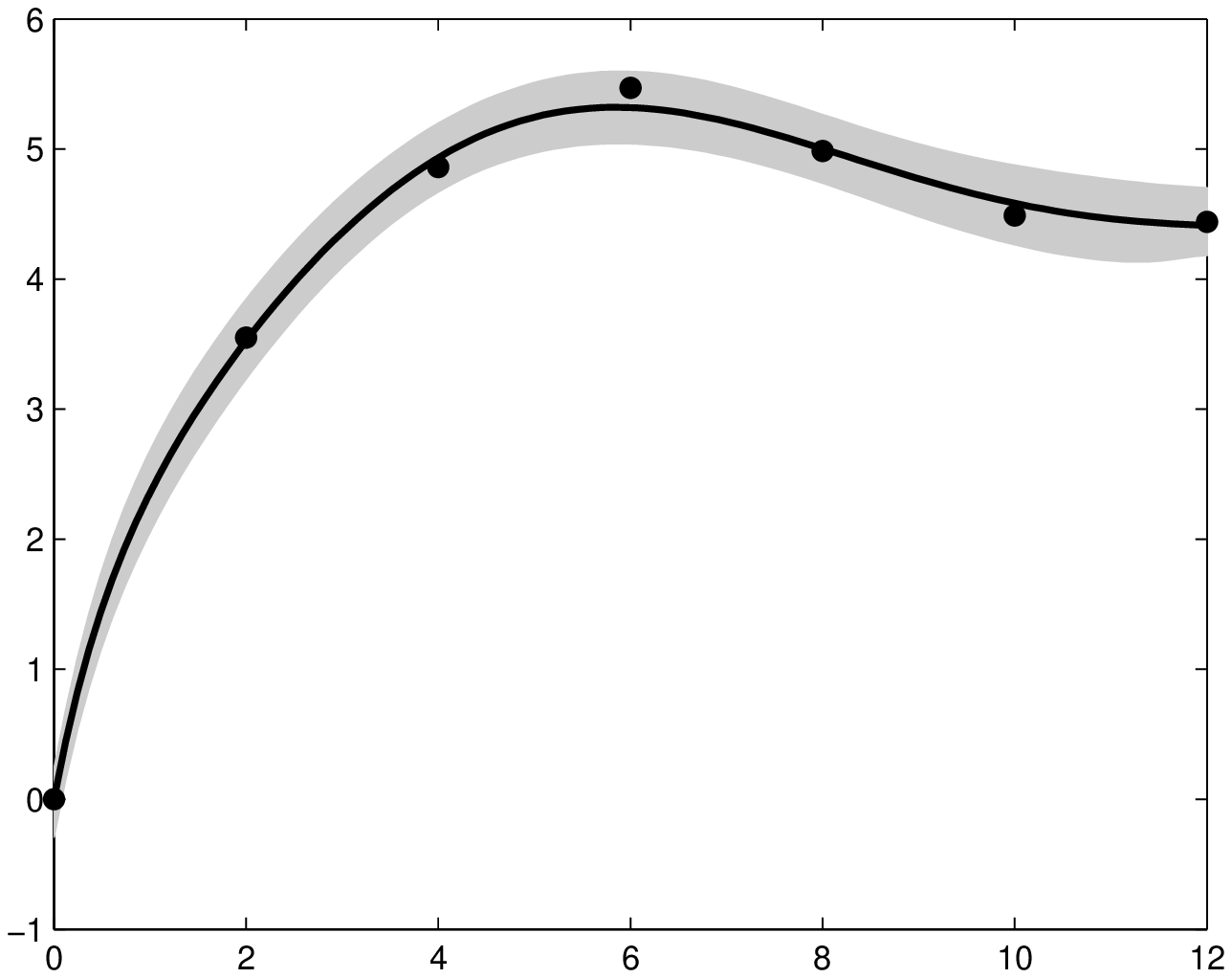} 
                \caption{BIK}
        \end{subfigure}
        \begin{subfigure}[b]{0.23\textwidth}
           \includegraphics[width=\textwidth]{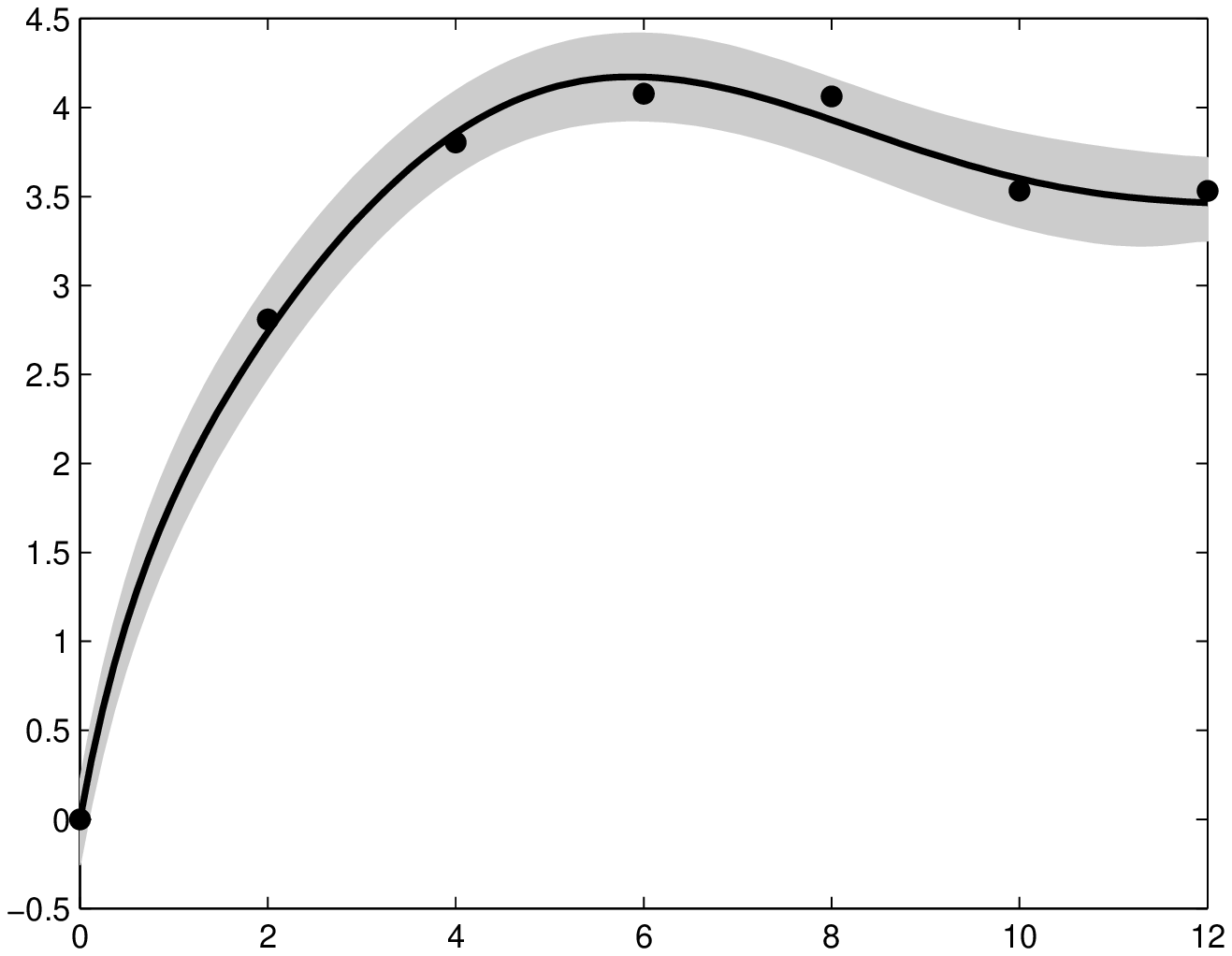} 
           \caption{TNFRSF10b}
        \end{subfigure}\\
        \begin{subfigure}[b]{0.23\textwidth}
           \includegraphics[width=\textwidth]{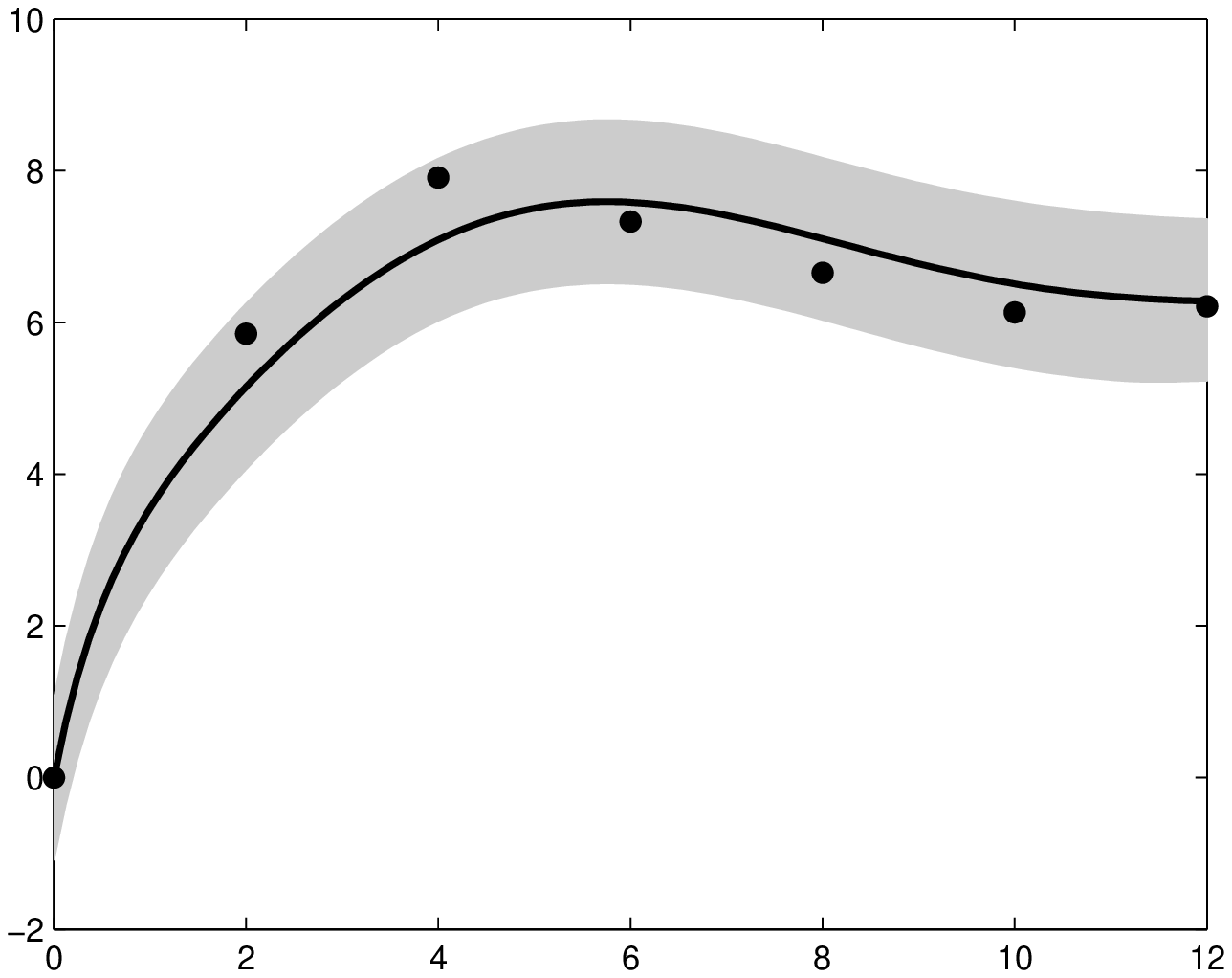} 
           \caption{CIp1/p21}
        \end{subfigure}
        \begin{subfigure}[b]{0.24\textwidth}
                \includegraphics[width=\textwidth]{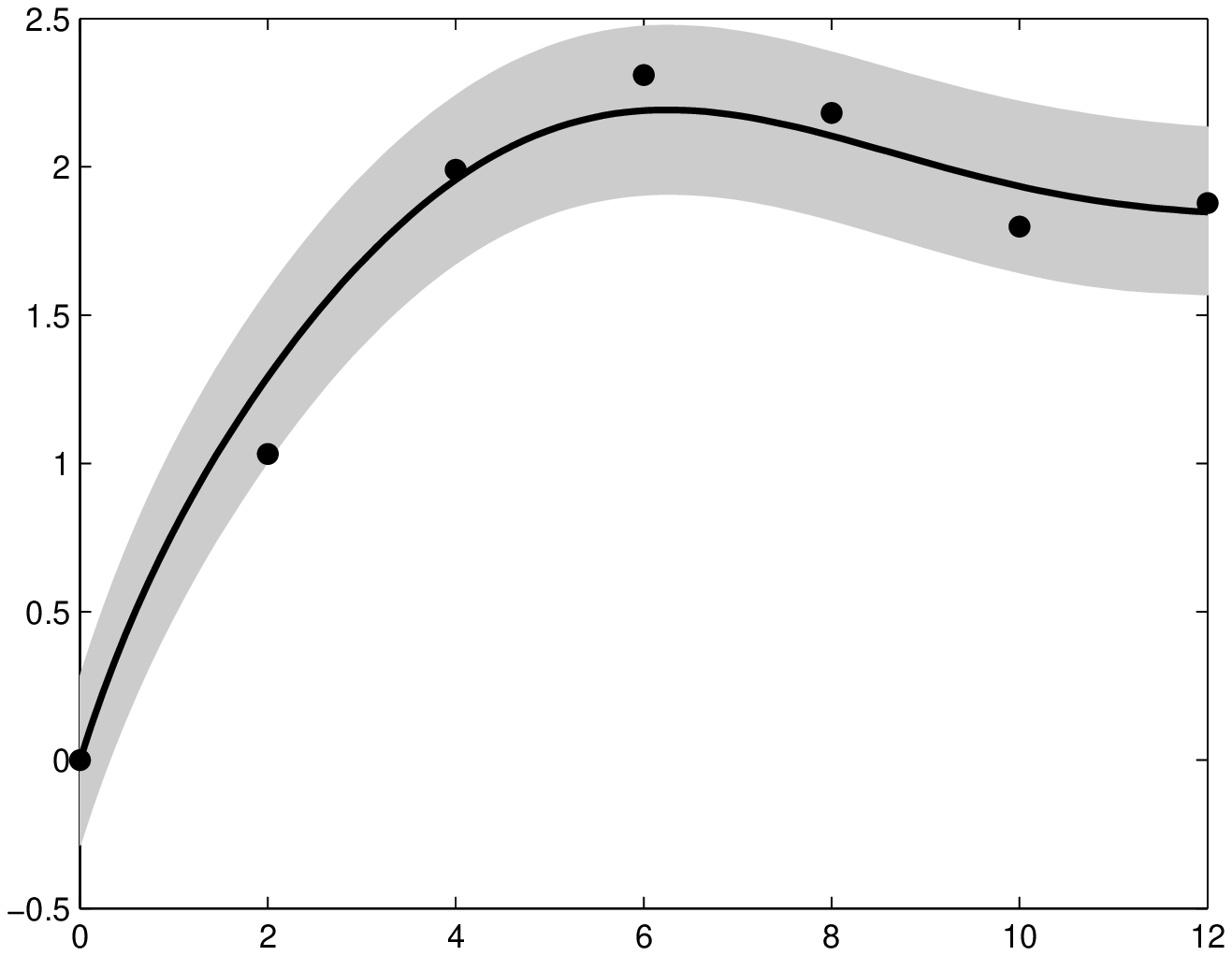} 
                \caption{SESN1/hPA26}
        \end{subfigure}%

        \caption{Mean (solid line) and two standard deviations (grey shade) for the predictions over five gene expression levels from the IBP + ODE1 model.}\label{fig:outputIBPODE1}
\end{figure}

\subsection{Gene expression data: Tumour Suppressor Protein p53}
Gene expression data consist of measurements of the mRNA concentration
of a set of genes. mRNA concentration for each gene is regulated by
the so called transcription factor
(TFs) proteins. In transcriptional regulatory networks, a TF or a set
of TFs may act in a individually or collaborative manner, leading to
complex regulatory interactions.

Gene expression data can be related to a first order differential equation
\citep{Barenco:2006}, with the same form given in
equation \ref{fos}. From this equation, and in the context of gene
expression, the output $f_d(t)$ is the mRNA concentration of gene $d$,
$B_d$ is the linear degradation rate of $f_d(t)$, and $u(t)$ is the
concentration of the TF. Two major problems arise from the analysis of
gene expression, first to determine the interaction network and second
to infer the activated transcription factor \citep{Gao:2008}.

In this experiment, we use tumour suppressor protein p53 dataset from
\cite{Barenco:2006}. This dataset is restricted to five known target
genes: DDB2, BIK, TNFRSF10b, CIp1/p21 and SESN1/hPA26. In
\cite{Lawrence:NIPS:2006}, a transcription factor is inferred from the
expression levels of these five target genes using a covariance function
build from a first order differential equation. Our aim is to
determine the number of transcription factors (latent functions) and
how they explain the activities of the target genes using the
covariance function defined in \ref{sec:ode1}.

\begin{figure}[ht!]
        \centering
                \includegraphics[width=0.22\textwidth, height=6cm]{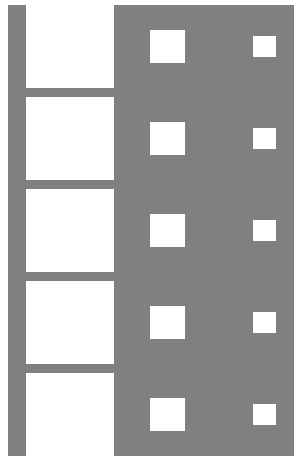}
                \label{fig:hbar}
        \caption{Hinton diagrams for the IBP variational approximation using ODE1 covariance function.}\label{fig:hintonbar}
\end{figure}

Results obtained from this dataset regarding the number of latent
forces, concurred with the description given in \citet{Barenco:2006},
where there is one protein influencing the expression level of the
genes analysed (see Figure \ref{fig:hintonbar}).

\section{Conclusions}

We have introduced a new variational method to perform model selection
in convolved multiple output Gaussian Processes. Our main aim was to
identify the relationship between the latent functions and the outputs
in multiple output Gaussian processes. The proposed method achieved
comparable results to the model that assumes full connectivity between
latent functions and output functions. This makes our
method suitable to applications where the complexity of the model
should be reduced.  The proposed model selection method can be applied
in other applications that involve the use of a covariance function
based on differential equations, such as inferring the biological
network in gene expression microarray data.

For the artificial dataset examples we found that the model selection
method converges to a similar matrix for the interconnections between
latent functions and outputs. We have illustrated the performance of
the proposed methodology for regression of human motion capture data,
and a small gene expression dataset.

\subsubsection*{Acknowledgments}
CDG would like to thank to Convocatoria 567 of Colciencias. MAA would
like to thank to Banco Santander for the support received under the
program ``Scholarship for Young Professors and Researchers
Iberoam\'erica''. MAA would also like to acknowledge the support from
British Council and Colciencias under the research project ``Sparse
Latent Force Models for Reverse Engineering of Multiple Transcription
Factors''. The authors of this manuscript would like to thank to
Professor Fernando Quintana from Pontificia Universidad Cat\'olica de
Chile, for his valuable discussions and feedback on this manuscript.

\bibliographystyle{plainnat}
\bibliography{ibpmultigp}

\appendix
\section{Computing the optimal posterior distributions}\label{UpDist}
In this appendix, we present the updates of variational distributions $q(\boldS,\boldZ)$, $q(\boldu)$, $q(\gamma)$ and $q(\upsilon)$. To do so, first, we rewrite the lower bound defined in section \ref{model}, as
\begin{align*}
F_{V} &= \sum_{q=1}^Q\tr\left(\mathbf{m}_q\ex[\boldu_q^{\top}]\right)
-\frac{1}{2}\sum_{q=1}^Q\sum_{q'=1}^Q\tr\left(\mathbf{P}_{q,q'}\ex[\mathbf{u}_{q'}\mathbf{u}^{\top}_{q}]\right)
-\frac{1}{2}\sum_{q=1}^Q\tr\left(\boldK_{\boldu_q,\boldu_q}^{-1}\ex[\boldu_q\boldu_q^{\top}]\right)-\frac{QM}{2}\log 2\pi\\
&-\frac{1}{2}\sum_{q=1}^Q\log |\boldK_{\boldu_q,\boldu_q}|+\entropy(\boldu)
-\frac{ND}{2}\log 2\pi -\frac{1}{2}\sum_{d=1}^D\log |\bm{\Sigma}_{\mathbf{w}_d}|-\frac{1}{2}\sum_{d=1}^D\tr\left
(\bm{\Sigma}_{\mathbf{w}_d}^{-1}\boldy_d\boldy_d^{\top}\right)
-\frac{1}{2}\log 2\pi\sum_{d=1}^D\sum_{q=1}^Q\ex[Z_{d,q}] \\
&+\sum_{d=1}^D\sum_{q=1}^Q \ex[Z_{d,q}]\ex[\log \pi_q] +
\frac{1}{2}\sum_{d=1}^D\sum_{q=1}^Q\ex[Z_{d,q}]\left[\psi(a^{\gamma^*}_{d,q})-\log b^{\gamma^*}_{d,q}\right]
-\frac{1}{2}\sum_{d=1}^D\sum_{q=1}^Q\left(\frac{a^{\gamma^*}_{d,q}}{b^{\gamma^*}_{d,q}}+c_{d,q}\right)\ex[Z_{d,q}S_{d,q}^2] \\
&+ \sum_{d=1}^D\sum_{q=1}^Q(1-\ex[Z_{d,q}])\ex[\log(1- \pi_q)]
-\sum_{d=1}^D\sum_{q=1}^Q\log\Gamma(a^{\gamma}_{d,q}) + \sum_{d=1}^D\sum_{q=1}^Q a^{\gamma}_{d,q}\log b^{\gamma}_{d,q}\\
&+ \sum_{d=1}^D\sum_{q=1}^Q\left(a^{\gamma}_{d,q}-1\right)\left[\psi(a^{\gamma^*}_{d,q}) -\log b^{\gamma^*}_{d,q}\right]
-\sum_{d=1}^D\sum_{q=1}^Q b^{\gamma}_{d,q}\frac{a^{\gamma^*}_{d,q}}{b^{\gamma^*}_{d,q}}
+ (\alpha - 1) \sum_{q=1}^Q\left[ \psi(\tau_{q1}) - \psi(\tau_{q1} + \tau_{q2}) \right] + Q\log \alpha\\
&+ \entropy(\boldS, \boldZ)+\entropy(\boldupsi) + \entropy(\boldgamma),
\end{align*}
with
\begin{align*}
c_{d,q} = & \tr\left(\bm{\Sigma}_{\mathbf{w}_d}^{-1}\boldK_{\boldf_d|\boldu_q}\right), \\ \boldK_{\boldf_d|\boldu_q} = &\mathbf{K}^{(q)}_{\mathbf{f}_d\mathbf{f}_d}-
\mathbf{K}_{\mathbf{f}_d\mathbf{u}_q}\mathbf{K}^{-1}_{\mathbf{u}_q\mathbf{u}_q}
\mathbf{K}_{\mathbf{f}_d,\mathbf{u}_q}^{\top},\\
\mathbf{m}_q = & \sum_{d=1}^D\ex[Z_{d,q}S_{d,q}]\boldK^{-1}_{\boldu_q,\boldu_q}\mathbf{K}^{\top}_{\mathbf{f}_d,\mathbf{u}_q}
\bm{\Sigma}_{\mathbf{w}_d}^{-1}\boldy_d,\\
\mathbf{P}_{q,q'}=&\sum_{d=1}^D\ex\left[Z_{d,q}S_{d,q}Z_{d,q'}S_{d,q'}\right]
\mathbf{K}^{-1}_{\mathbf{u}_q,\mathbf{u}_q}\mathbf{K}_{\mathbf{f}_d,\mathbf{u}_q}^{\top}\bm{\Sigma}^{-1}_{\mathbf{w}_d}\mathbf{K}_{\mathbf{f}_d,\mathbf{u}_{q'}}\mathbf{K}^{-1}_{\mathbf{u}_{q'},\mathbf{u}_{q'}}.
\end{align*}

Additionally, $\bm{\Sigma}_{\mathbf{w}}$ is the covariance matrix for the
observation noise.

\subsection{Updates for distribution $q(\boldu)$}
Taking into account that $q(\boldu) = \prod\limits_{q=1}^{Q} q(\boldu_q)$ and $q(\boldu_q)= \mathcal{N} (\boldu_q | \widetilde{\boldu}_q,\widetilde{\boldK}_{\boldu_q,\boldu_q})$. Then, it can be shown that the moment updates are
\[   \widetilde{\boldK}_{\boldu_i,\boldu_i}^{-1} = \mathbf{P}_{i,i} + \boldK_{\boldu_i,\boldu_i}^{-1} \quad  \widetilde{\boldu}_q = \widetilde{\boldK}_{\boldu_i,\boldu_i} \widehat{\boldu}_i,
\]
with 
\[ \widehat{\boldu}_i = \sum_{d=1}^D\ex[Z_{d,i}S_{d,i}]\boldK^{-1}_{\boldu_i,\boldu_i}\mathbf{K}^{\top}_{\mathbf{f}_d,\mathbf{u}_i}
\bm{\Sigma}_{\mathbf{w}_d}^{-1} \left( \boldy_d - \widehat{\boldy}\right), \]
and
\[ \widehat{\boldy} = \sum_{q'=1,q'\neq i}^Q \ex\left[Z_{d,q'}S_{d,q'}\right]
\mathbf{K}_{\mathbf{f}_d,\mathbf{u}_{q'}}\mathbf{K}^{-1}_{\mathbf{u}_{q'},\mathbf{u}_{q'}}   \ex[\mathbf{u}_{q'}].
\]

\subsection{Updates for distribution $q(\boldS,\boldZ)$}
We assume that the distribution $q(\boldS, \boldZ)$ is defined as
\[ q(\boldS,\boldZ) = q(\boldS | \boldZ) q(\boldZ) = \prod\limits_{d=1}^{D}\prod\limits_{q=1}^{Q} q(S_{d,q}|Z_{d,q})q(Z_{d,q}).
\]
First, we calculate the update parameters for the distribution $q(Z_{d,q})$, which is defined as
\[ q(Z_{d,q}) = \eta_{d,q}^{Z_{d,q}}(1-\eta_{d,q})^{1-Z_{d,q}}.
\]

Also, $q(S_{d,q}|Z_{d,q})$ is defined as
\[ q(S_{d,q}|Z_{d,q}=1) = \mathcal{N} (S_{d,q}|\mu_{S_{d,q}}, \nu_{d,q}).
\]

Then it can be shown that the update for $\nu_{d,i}$ is given by
\begin{align*}
\nu_{d,i}^{*} &= \left(\tr\left( \mathbf{P}_{d,i,i} \left[ \widetilde{\boldK}_{\boldu_i,\boldu_i}+\ex[\boldu_i]\ex[\boldu_i]^{\top} \right] \right)
+\frac{a^{\gamma^*}_{d,i}}{b^{\gamma^*}_{d,i}}+c_{d,i} \right)^{-1},
\end{align*}
with 
\begin{align*}
\mathbf{P}_{d,q,q'}&=\mathbf{K}^{-1}_{\mathbf{u}_q,\mathbf{u}_q}\mathbf{K}_{\mathbf{f}_d,\mathbf{u}_q}^{\top}\bm{\Sigma}^{-1}_{\mathbf{w}_d}\mathbf{K}_{\mathbf{f}_d,\mathbf{u}_{q'}}\mathbf{K}^{-1}_{\mathbf{u}_{q'},\mathbf{u}_{q'}}.
\end{align*}
While, the update for parameter $\mu_{S_{d,i}}$ is calculated as
\begin{align*}
\mu_{S_{d,i}} &= \nu_{d,i}^* \left[ \tr\left(\mathbf{m}_{d,i}\ex[\boldu_i^{\top}]\right)
-\sum_{q'=1,q'\neq i}^Q\tr\left(\eta_{d,q'}\mu_{S_{d,q'}} \mathbf{P}_{d,i,q'}\ex[\mathbf{u}_{q'}]\ex[\mathbf{u}^{\top}_{i}]\right)
\right],
\end{align*}
with 
\begin{align*}
\mathbf{m}_{d,q} &= \boldK^{-1}_{\boldu_q,\boldu_q}\mathbf{K}^{\top}_{\mathbf{f}_d,\mathbf{u}_q}
\bm{\Sigma}_{\mathbf{w}_d}^{-1}\boldy_d.
\end{align*}
Finally, the update for $\eta_{d,i}$ is given by

\begin{align*}
\ln \frac{ \eta_{d,i} } {1-\eta_{d,i}} = \vartheta_{d,i} &= \tr\left(\mu_{S_{d,i}}\mathbf{m}_{d,i}\ex[\boldu_i^{\top}]\right)
- \sum_{q'=1,q'\neq i}^Q\tr\left( \mu_{S_{d,i}}\eta_{d,q'}\mu_{S_{d,q'}} \mathbf{P}_{d,i,q'}\ex[\mathbf{u}_{q'}\mathbf{u}^{\top}_{i}]\right) \\
&-\frac{1}{2} \tr\left( \left( \nu_{d,i} + \mu_{S_{d,i}}^2 \right) \mathbf{P}_{d,i,i} \left[ \widetilde{\boldK}_{\boldu_i,\boldu_i} + \ex[\boldu_i]\ex[\boldu_i]^{\top} \right] \right)
-\frac{1}{2}\log 2\pi
+ \ex[\log \pi_i] 
\\&+ \frac{1}{2}\left[\psi(a^{\gamma^*}_{d,i})-\log b^{\gamma^*}_{d,i}\right]
-\frac{1}{2} \left(\frac{a^{\gamma^*}_{d,i}}{b^{\gamma^*}_{d,i}}+c_{d,i}\right)\left( \nu_{d,i} + \mu_{S_{d,i}}^2 \right)
- \ex[\log(1- \pi_i)] + 
\frac{1}{2}\ln (2\pi e^{1} \nu_{d,i}),
\end{align*}
where
\[ \eta_{d,i} = \frac{1}{1+e^{-\vartheta_{d,i}}}, \quad \ex_{q(\boldupsi)}[\log\pi_q] = \sum_{i=1}^q [\psi(\tau_{i1}) - \psi(\tau_{i1} + \tau_{i2})].
\]
For computing $\ex_{q(\boldupsi)}[\log (1-\prod_{i=1}^q\upsilon_i)]$, we would need to resort to a local variational approximation \citep{Bishop:PRLM06}  in a similar way to \citet{DoshiVelez:VIBP:2009}.
\subsection{Updates for distribution $q(\boldgamma)$}
This distribution is defined as
\[ q(\boldgamma) = \prod_{d=1}^{D}\prod_{q=1}^{Q} \text{Gamma} (\gamma_{d,q}|a_{d,q}^{\gamma^*},b_{d,q}^{\gamma^*}).
\]
It can be shown that the updates for the parameters $b^{\gamma^*}_{d,q}$ and $ a^{\gamma^*}_{d,q}$ are given by
\[ b_{d,q}^{\gamma^*} = \frac{1}{2}  \ex[Z_{d,q}S_{d,q}^2] + b^{\gamma}_{d,q},
\]
and
\[ a^{\gamma^*}_{d,q} =\frac{1}{2}\ex[Z_{d,q}] + a^{\gamma}_{d,q}.
\]

\subsection{Updates for distribution $q(\bm{\upsilon})$}
We assumed that the optimal distribution for each $q(\upsilon_i) = \betad(\upsilon_i|\tau_{i1}, \tau_{i2})$. Given a fixed
value for $i$, the updates for parameters $\tau_{i1}$ and $\tau_{i2}$ are given by
\begin{align*}
\tau_{i1} & = \alpha + \sum_{m=i}^Q\sum_{d=1}^D \eta_{d,m} + \sum_{m=i+1}^Q\left[\sum_{d=1}^D(1 - \eta_{d,m})\sum_{j=i+1}^m
q_{mj}\right],\\
\tau_{i2} & = 1 +  \sum_{m=i}^Q\sum_{d=1}^D (1 - \eta_{d,m})q_{mi}.
\end{align*}

\section{Predictive distribution}\label{PreDist}

For the predictive distribution, we need to compute
\begin{align*}
p(\boldy_*|\bm{\theta})&=\int_{\boldS, \boldZ}\int_{\boldu,u}
p(\boldy_*|\bm{\theta}, u,\boldS, \boldZ)q(u,\boldu)
q(\boldS, \boldZ)\dif{u}\;\dif{\boldu}\;\dif{\boldS}\;\dif{\boldZ}.
\end{align*}

It can be demonstrated that the above integral is intractable. Since we are only interested in the mean and covariance
for $\boldy_*$, we can still compute them using
\begin{align*}
\ex[\boldy_*]&=\int_{\boldy_*}\boldy_{*}p(\boldy_*|\cdot)\dif{\boldy_{*}}=
\int_{\boldS, \boldZ}\left[\int_{\boldy_*}\boldy_{*}p(\boldy_*|\cdot)\dif{\boldy_{*}}\right]
q(\boldS, \boldZ)\dif{\boldS}\;\dif{\boldZ}
=\int_{\boldS, \boldZ}\ex_{\boldy_*|\cdot}[\boldy_*]q(\boldS, \boldZ)\dif{\boldS}\;\dif{\boldZ},
\end{align*}
we get
\begin{align*}
\ex_{\boldy_d}[\boldy_d]&=\sum_{k=1}^Q\ex[Z_{d,k}S_{d,k}] \widehat{\bm{\alpha}}_{d,k}
\end{align*}
where $\widehat{\bm{\alpha}}_{d,k}=\boldK_{\boldf_d,\boldu_k}\boldA^{-1}_{k,k}\left(\sum_{j=1}^D
\widetilde{\boldK}_{\boldu_k,\boldf_j}\bm{\Sigma}^{-1}_{\mathbf{w}_j}\widetilde{\boldy}_j\right)$, and
$\widehat{\bm{\alpha}}_{d} = [\widehat{\bm{\alpha}}_{d,1}, \ldots,
\widehat{\bm{\alpha}}_{d,Q}]$. Notice that the $\boldK_{\boldf_d, \boldu_k}$ in the expression for
$\widehat{\bm{\alpha}}_{d,k}$ must be computed at the test point $\boldx_*$. \\

We need to compute now the second moment of $\boldy_*$ under $p(\boldy_*|\cdot)$. Again, we can write
\begin{align*}
\ex[\boldy_*\boldy_*^{\top}]&=\int_{\boldy_*}\boldy_*\boldy_*^{\top}p(\boldy_*|\cdot)\dif{\boldy_{*}}=
\int_{\boldS, \boldZ}
\left[\int_{\boldy_*}\boldy_*\boldy_*^{\top}p(\boldy_*|\cdot)\dif{\boldy_{*}}\right]
q(\boldS, \boldZ)\dif{\boldS}\;\dif{\boldZ}=
\int_{\boldS, \boldZ}\ex_{\boldy_*|\cdot}[\boldy_*\boldy_*^{\top}]q(\boldS, \boldZ)\dif{\boldS}\;\dif{\boldZ},
\end{align*}
where $\ex_{\boldy_*|\cdot}[\boldy_*\boldy_*^{\top}]$ is the second moment of $\boldy_*$ under the density
$p(\boldy_*|\cdot)$,
we get
\begin{align*}
\ex_{\boldy_*}[\boldy_d\boldy_d^{\top}]&=\sum_{i=1}^Q\ex[Z_{d,i}S_{d,i}^2]\boldK^{(i)}_{\boldf_d,\boldf_{d}}
-\sum_{i=1}^{Q}\ex[Z_{d,i}S_{d,i}^2]\boldK_{\boldf_d,\boldu_i}\bm{\Gamma}_{i,i}\boldK_{\boldu_i,\boldf_{d}}
-\sum_{i=1}^{Q}\sum_{j=1}^{Q}\ex[Z_{d,i}S_{d,i}Z_{d,j}S_{d,j}]\widehat{\bm{\alpha}}_{d,i}\widehat{\bm{\alpha}}^{\top}_{d,j}
+ \bm{\Sigma_{\mathbf{w}_d}}.
\end{align*}
Finally, the covariance $\cov[\boldy_d\boldy_d^{\top}]$ would be given as
\begin{align*}
\cov[\boldy_d\boldy_d^{\top}]&=
\sum_{i=1}^Q\ex[Z_{d,i}S_{d,i}^2]\boldK^{(i)}_{\boldf_d,\boldf_{d}}
-\sum_{i=1}^{Q}\ex[Z_{d,i}S_{d,i}^2]\boldK_{\boldf_d,\boldu_i}\bm{\Gamma}_{i,i}\boldK_{\boldu_i,\boldf_{d}}+ \bm{\Sigma_{\mathbf{w}_d}}.
\end{align*}
Bear in mind, we have omitted $*$ in the vectors above to keep the notation uncluttered.

\end{document}